%% file: main.tex
\documentclass{l4dc2027}
\usepackage{times}

\usepackage{amsfonts}
\usepackage{textcomp}
\usepackage[dvipsnames]{xcolor}
\usepackage{booktabs}
\usepackage{tikz}
\usetikzlibrary{arrows.meta,positioning}
\hypersetup{pdftitle={Specification-Aware Sparse Differentiation: Planning under Signal Temporal Logic with Sparsemax Robustness}}
\usepackage[capitalise,noabbrev]{cleveref}
\usepackage{macros}

\graphicspath{{figures/}}

\title[Specification-Aware Sparse Differentiation]{Specification-Aware Sparse Differentiation: Planning under Signal Temporal Logic with Sparsemax Robustness\titletag{\thanks{Code available at \noexpand\url{https://github.com/sdamera95/SparseMax-dSTL}.}}}

\author{%
 \Name{Sai Sandeep Damera} \Email{sdamera@umd.edu}\\
 \Name{Ryan Matheu} \Email{rmatheu@umd.edu}\\
 \Name{John S. Baras} \Email{baras@umd.edu}\\
 \Name{Calin Belta} \Email{calin@umd.edu}\\
 \addr University of Maryland, College Park, USA%
}

\begin{document}

\maketitle

\begin{abstract}
Gradient-based optimal control under a Signal Temporal Logic (STL) specification involves solving a nonlinear program whose constraints are smoothed robustness functions of the trajectory, and automatic differentiation without knowledge of sparsity computes the Jacobian and the Hessian that its solver needs in full. This often renders the problem computationally intractable. The dynamics and the specification prescribe a sparsity pattern through the STL syntax, the windows of the temporal operators and the variables that the predicates read. Standard tools such as CasADi recover this pattern by structural detection. In this work, we propose a method to prescribe such a pattern directly from the task specification. We show that our method allows us to exploit SparseMax, a recently proposed smooth robustness operator that is certifiably sparse and sound. Its derivative is exactly zero at every operand at or below a data-dependent threshold, which structural detection cannot see. We derive the patterns that expose this trajectory-dependent sparsity to sparse automatic differentiation, with a margin test for their reuse and Hessians confined to the variables of each constraint's support. We demonstrate the method on a dual-arm coordination task in which two arms share a hand-off zone divided into cells and one until constraint per scheduled visit keeps the other arm out of a cell until the visitor has left it. The two scenes are a zone of five cells and a circle of 64 cells, with single solves and closed-loop re-planning on a rigid-body simulation. Our exact-Hessian interior-point solves are up to 27 times faster than those with the sound log-sum-exp robustness in CasADi and IPOPT.
\end{abstract}


\input{sections/01-introduction}

\input{sections/02-related}
\input{sections/03-preliminaries}
\input{sections/04-patterns}
\input{sections/06-second-order}
\input{sections/08-plan}
\input{sections/09-results}

\input{sections/10-conclusion}

\acks{The authors acknowledge the use of Claude LLMs for code development, visualizations, and man\-u\-script review.}

\bibliography{references}
\newpage
\appendix
\crefalias{section}{appendix}
\crefalias{subsection}{subappendix}
\crefalias{subsubsection}{subsubappendix}
\input{sections/A-appendix}

\end{document}

%% file: sections/01-introduction.tex
\section{Introduction}
\label{sad:sec:introduction}
Consider two arms that pass through a shared hand-off zone. The zone is divided into cells, only one arm may occupy a cell at a time, and each arm visits the cells on a schedule and must reach waypoints on time. A Signal Temporal Logic (STL) specification combines predicates on the state, such as an arm being outside a cell, with Boolean and temporal operators, and its robustness is positive when a trajectory satisfies it. Ours is a conjunction whose operands, the conjuncts, are one until per scheduled visit, which keeps the other arm out of the cell until the visitor has left it within a time window, and one eventually per waypoint. A receding-horizon controller enforces it by solving, at every control period, a nonlinear program over the next $T$ time steps subject to the dynamics and to a smoothed robustness of every conjunct at or above a margin. An interior-point solver needs the constraint Jacobian at every update and, for exact second-order steps, the Hessian of the Lagrangian. Automatic differentiation without knowledge of sparsity computes both in full. At 400 time steps that is 11{,}222 reverse and 16{,}800 Hessian-vector products at every update.

\begin{figure}[t]
\centering
\resizebox{\textwidth}{!}{\input{figures/framework-overview}}
\caption{Overview: the dual-arm task with one until constraint, the prescribed sparsity pattern, its restriction to the sparsemax supports at the iterate, and the test for its reuse.}
\label{sad:fig:overview}
\end{figure}
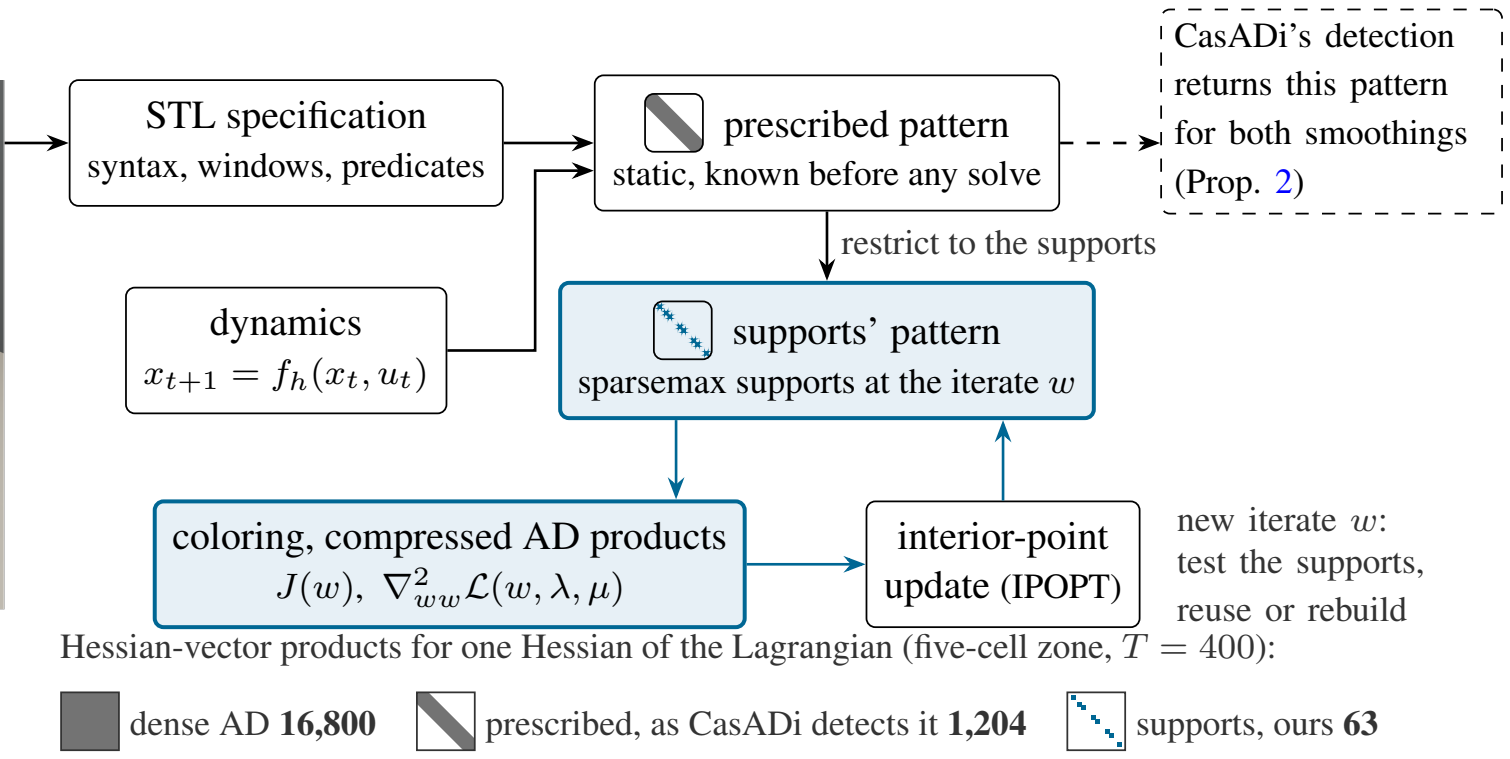

Most entries of these two matrices are zero, for reasons known before any derivative is evaluated. A defect constraint of the dynamics reads two consecutive time steps, and a conjunct reads the time steps inside the windows of its temporal operators and, at each of them, only the coordinates on which its predicates depend. The specification and the dynamics therefore prescribe a sparsity pattern, and a coloring of it lets sparse automatic differentiation recover every entry from a few compressed products \citep{curtis1974estimation,gebremedhin2005color,gebremedhin2007new}. CasADi detects this structure from the program itself. It propagates dependencies through the expression graph of the program, colors the detected pattern and hands sparse derivatives to the solver \citep{andersson2019casadi,wachter2006implementation}, which is one reason it is the tool of choice for such problems.

How sparse the derivatives are inside this pattern depends on the smoothing. The log-sum-exp robustness \citep{pant2017smooth,gilpin2020smooth} can be positive on a violating trajectory in its common form, and its sound variant (\slse) \citep{mehdipour2025generalized,damera2026sparsemax}, our baseline, gives every operand of every extremum a positive weight. Sparsemax robustness \citep{damera2026sparsemax} is a lower bound of the exact robustness that gives a derivative of exactly zero to every operand at or below a data-dependent threshold. We show what the set of operands with a nonzero weight, the support, gives computationally. First, structural detection, which propagates dependencies through the operations of a program as CasADi does, returns the same pattern for sparsemax as for log-sum-exp (\cref{sad:prop:detector}). Second, a method that evaluates the constraints once, reads the supports and restricts the prescribed pattern to them obtains a smaller pattern at that trajectory, with a margin that states when it remains valid at the next one (\cref{sad:sec:patterns,sad:fig:overview}).

The gain is largest at second order. The Hessian pattern of a log-sum-exp conjunct is dense over every variable in its windows, while that of a sparsemax conjunct is confined to the variables of its support (\cref{sad:sec:second-order}). On the task above at 400 time steps the Hessian of the Lagrangian takes 1{,}204 Hessian-vector products with the prescribed pattern and 63 with the supports, and on a circle of 64 cells the constraint Jacobian takes 102 products against 13. With exact Hessians on the supports' structure, a solve from the previous period's solution at 200 time steps takes 3 updates of the iterate and 0.22~s, 18 times shorter than log-sum-exp in CasADi with IPOPT, whose exact-Hessian solve takes the same 3 updates. Re-planning every 0.5~s after one arm is held up, our controller completes each control period over a horizon of 2~s within 0.40~s, inside the control period.

We contribute (i) the sparsity pattern of the constraint Jacobian that the dynamics and an STL specification prescribe, with the proof that every gradient lies inside it (\cref{sad:sec:patterns}); (ii) \cref{sad:prop:detector} on what structural detection returns for sparsemax and for log-sum-exp, with a radius within which the pattern of the current supports, and a cached coloring, stay valid; (iii) the sparse exact Hessian, supported on the square of the active pattern of each constraint (\cref{sad:sec:second-order}); and (iv) a comparison with dense differentiation and with log-sum-exp in CasADi and IPOPT on a dual-arm task, in products, solve times and the control period of a re-planning loop (\cref{sad:sec:plan}).

%% file: figures/framework-overview.tex
\begin{tikzpicture}[
  font=\footnotesize,
  box/.style={draw, rounded corners=2pt, align=center, inner sep=4pt, minimum height=10mm, fill=white},
  ours/.style={box, draw=MidnightBlue, line width=0.9pt, fill=MidnightBlue!6},
  note/.style={align=left, inner sep=1pt, text=black!80},
  arr/.style={-{Stealth[length=5pt]}, semithick},
  oarr/.style={-{Stealth[length=5pt]}, semithick, MidnightBlue}]
\newcommand{\glyphdense}{\tikz[baseline=1.2mm]{\fill[black!55] (0,0) rectangle (0.46,0.46);\draw (0,0) rectangle (0.46,0.46);}}
\newcommand{\glyphband}{\tikz[baseline=1.2mm]{\begin{scope}\clip (0,0) rectangle (0.46,0.46);\draw[black!55, line width=1.3mm] (-0.05,0.51) to (0.51,-0.05);\end{scope}\draw (0,0) rectangle (0.46,0.46);}}
\newcommand{\glyphsupp}{\tikz[baseline=1.2mm]{\foreach \x/\y in {0.03/0.40,0.08/0.35,0.11/0.32,0.19/0.24,0.22/0.21,0.30/0.13,0.33/0.10,0.40/0.03} \fill[MidnightBlue] (\x,\y) rectangle ++(0.035,0.035);\draw (0,0) rectangle (0.46,0.46);}}
\node[inner sep=0pt, anchor=north west, draw=black!40, line width=0.3pt] (img) at (-1.65,2.45) {\IfFileExists{figures/overview-render.png}{\includegraphics[width=6.3cm]{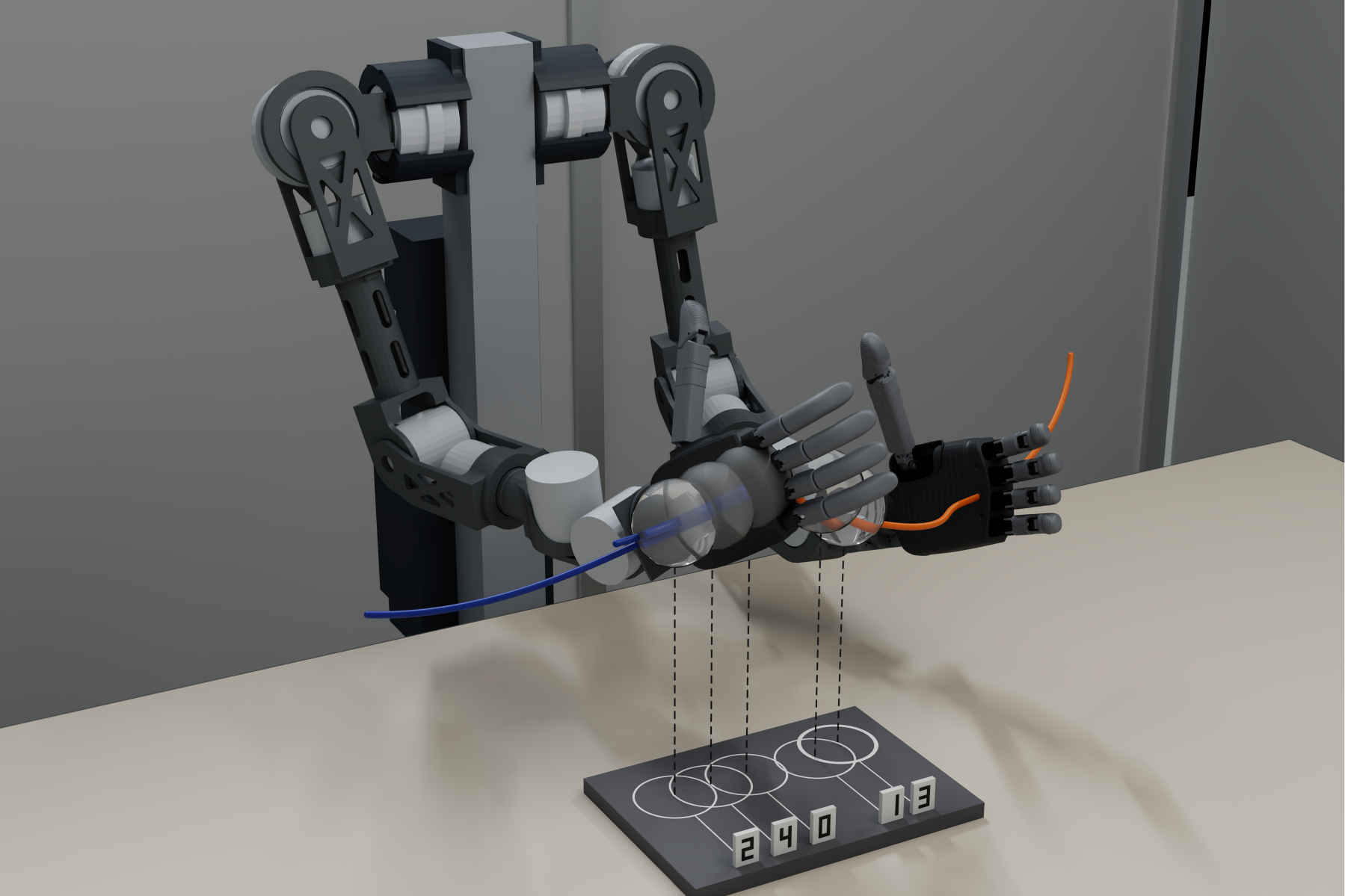}}{\textcolor{black!12}{\rule{6.3cm}{4.2cm}}}};
\node[note, anchor=north west, text width=6.3cm] (cap) at ([yshift=-1.5mm]img.south west) {{\scriptsize The right arm inside cell 3 while the left arm stays out: $\mathrm{out}_{\mathrm{L},3}\,\mathbf{U}_{[0,48]}\,\mathrm{out}_{\mathrm{R},3}$}};
\node[box] (spec) at (6.9,1.95) {STL specification\\{\scriptsize syntax, windows, predicates}};
\node[box] (dyn) at (6.9,0.3) {dynamics\\{\scriptsize $x_{t+1}=f_h(x_t,u_t)$}};
\node[box] (stat) at (11.2,1.95) {\glyphband\ \ prescribed pattern\\{\scriptsize static, known before any solve}};
\node[ours] (act) at (11.2,0.3) {\glyphsupp\ \ supports' pattern\\{\scriptsize sparsemax supports at the iterate $w$}};
\node[box, dashed, text width=2.5cm, align=left, inner sep=3pt, anchor=west] (det) at (13.85,2.2) {{\scriptsize CasADi's detection returns this pattern for both smoothings (Prop.~\ref{sad:prop:detector})}};
\node[ours] (ad) at (8.2,-1.4) {coloring, compressed AD products\\{\scriptsize $J(w)$,\ \ $\nabla^2_{ww}\mathcal L(w,\lambda,\mu)$}};
\node[box] (ip) at (12.6,-1.4) {interior-point\\update {\scriptsize (IPOPT)}};
\node[note, anchor=west, text width=2.45cm] (loop) at ([xshift=2.5mm]ip.east) {{\scriptsize new iterate $w$:\\test the supports,\\reuse or rebuild}};
\draw[arr] (img.east |- spec) to (spec);
\draw[arr] (spec) to (stat);
\draw[arr] (dyn.east) to ([xshift=2.5mm]spec.east |- dyn.east) |- ([yshift=-2.2mm]stat.west);
\draw[arr] (stat) to node[right, note, xshift=2pt] {\scriptsize restrict to the supports} (act);
\draw[arr, dashed] (stat.east) to (stat.east -| det.west);
\draw[oarr] ([xshift=-12mm]act.south) to ([xshift=-12mm]act.south |- ad.north);
\draw[oarr] (ad) to (ip);
\draw[oarr] (ip.north) to (ip.north |- act.south);
\node[note, anchor=north west] at ([xshift=4mm]img.south east |- cap.north) {{\scriptsize Hessian-vector products for one Hessian of the Lagrangian (five-cell zone, $T=400$):}};
\node[note, anchor=north west] at ([xshift=4mm, yshift=-4.6mm]img.south east |- cap.north) {\glyphdense\ {\scriptsize dense AD \textbf{16{,}800}}\quad\glyphband\ {\scriptsize prescribed, as CasADi detects it \textbf{1{,}204}}\quad\glyphsupp\ {\scriptsize supports, ours \textbf{63}}};
\end{tikzpicture}

%% file: sections/02-related.tex
\label{sad:sec:related}\textbf{Related work.} STL robustness reduces Boolean and temporal operators to nested extrema over predicate values \citep{maler2004monitoring,donze2010robust}, and smooth measures replace these extrema so that gradient-based solvers can enforce a specification \citep{pant2017smooth,leung2023backpropagation,kapoor2025stlcg++}. A sound measure pairs under-approximations of both extrema \citep{gilpin2020smooth}, and average-based measures form a further family \citep{mehdipour2019arithmetic,uzun2024optimization}, among them the generalized-mean robustness (GMR) \citep{mehdipour2025generalized}. An exact reformulation of the extrema removes the smoothing error by adding variables and constraints \citep{han2025exact}. Sparsemax robustness \citep{damera2026sparsemax}, on which we build, bounds the exact robustness from below and confines the gradient of each extremum to a data-dependent support. We derive the sparsity of its first and second derivatives from that support and use it in receding-horizon control, where STL specifications have been enforced \citep{raman2014model,sadraddini2015robust} and the cost of the derivatives recurs every control period.

Sparse automatic differentiation recovers a Jacobian or a Hessian from one product per color of a known sparsity pattern \citep{curtis1974estimation,gebremedhin2005color,gebremedhin2007new}. Structural detection of the pattern, as in CasADi, propagates dependencies through the operations and returns a pattern that is valid at every input \citep{andersson2019casadi,hill2025sparser}. Local detection also evaluates the operations and returns the pattern at one input \citep{hill2025sparser}. Our patterns are read from the supports in the forward evaluation of the constraints, without tracing the program, and they carry a margin that states at which other inputs they remain valid, together with the structure of the Hessian and its assembly in closed form. \Cref{sad:app:related} extends this survey.

%% file: sections/03-preliminaries.tex
\section{Preliminaries}
\label{sad:sec:preliminaries}

\subsection{Sampled STL robustness and sparsemax lower extrema}
\label{sad:subsec:lower}
Let $x_{t+1}=f_h(x_t,u_t)$ be a discrete-time robot model with sampling period $h>0$, state $x_t\in\R^{n_x}$ and control $u_t\in\R^{n_u}$, over a horizon of $T$ time steps. An STL specification in negation normal form is built from the grammar
\begin{equation}
\phi::=\mu\mid\neg\mu\mid\phi\wedge\psi\mid\phi\vee\psi\mid\mathbf G_I\phi\mid\mathbf F_I\phi\mid\phi\,\mathbf U_I\,\psi,
\label{sad:eq:grammar}
\end{equation}
where a predicate $\mu$ holds when $g_\mu(x_t)\ge0$, with $\rho^\mu_t=g_\mu(x_t)$ and $\rho^{\neg\mu}_t=-g_\mu(x_t)$, and $I=\{a,\ldots,b\}$ with integers $0\le a\le b$ is a window of sample offsets. The robustness at sample $t$ \citep{maler2004monitoring,donze2010robust} is defined recursively as the minimum over the operands for $\wedge$ and over the window for $\mathbf G_I$, the maximum for $\vee$ and $\mathbf F_I$, and for the until
\begin{equation}
\rho^{\phi\,\mathbf U_I\,\psi}_t=\max_{j\in I}\,\min\Bigl\{\rho^\psi_{t+j},\ \min_{0\le i\le j}\rho^\phi_{t+i}\Bigr\},
\label{sad:eq:semantics}
\end{equation}
with a closed prefix \citep{sadraddini2015robust}, and the sampled trajectory satisfies $\phi$ when $\rho^\phi_0\ge0$. Unrolling \cref{sad:eq:semantics} over the samples gives a finite acyclic evaluation graph of minima and maxima over predicate values, where a node $v$ has $m_v$ children $\ch(v)$ (\cref{sad:app:prelim}).

In our earlier work \citep{damera2026sparsemax}, we proposed the \smax{} operator and proved its soundness and sparsity by construction. In this work, we exploit the sparsity that arises from allocating the gradient signal on the support to boost the computational performance of optimal control under STL specifications. For $z\in\R^m$ and $\gamma>0$, the quadratically regularized maximum $M_\gamma(z)=\max_{p\in\mathcal S_m}\bigl(p^\top z-\tfrac\gamma2\|p\|_2^2\bigr)$ over the simplex $\mathcal S_m$ has the gradient $p_\gamma(z)=\Pi_{\mathcal S_m}(z/\gamma)$, the sparsemax projection \citep{martins2016Sparsemax}. The projection and the lower extrema are
\begin{align}
p_{\gamma,i}(z)&=\frac{(z_i-\theta)_+}{\gamma},\qquad \sum_{i=1}^{m}(z_i-\theta)_+=\gamma,\label{sad:eq:threshold}\\
Q^+_\gamma(z)&=M_\gamma(z)+\frac\gamma{2m},\qquad Q^-_\gamma(z)=-M_\gamma(-z)-\frac\gamma2.\label{sad:eq:lower}
\end{align}
By \cref{sad:eq:threshold}, every entry at or below the threshold $\theta$ receives a derivative of exactly zero. Both operators of \cref{sad:eq:lower} are continuously differentiable and nondecreasing in every entry, with gradients $p_\gamma(z)$ and $p_\gamma(-z)$ in the simplex, and each lies below its exact extremum by at most the band $b_m(\gamma)=\tfrac\gamma2(1-1/m)$ \citep[Lemma~1]{mensch2018differentiable}. We require $f_h$ and each $g_\mu$ to be continuously differentiable, twice where second derivatives are used.

Replace the extremum at every internal node $v$ of the evaluation graph by the operator of \cref{sad:eq:lower} with $m_v$ entries and parameter $\gamma_v$, and write $y_v$ or $\widetilde\rho_v$ for the smoothed value of the node. The smoothed value lies below the exact one by at most a budget that is known before optimization (\cref{sad:thm:lower}, from that paper), so a smoothed robustness of at least a margin $\delta\ge0$ implies an exact sampled robustness of at least $\delta$. By the chain rule $\nabla y_v=\sum_{c\in\ch(v)}\alpha_{v,c}\nabla y_c$, with the edge weights
\begin{equation}
\alpha_{v}=p_{\gamma_v}(s_vz_{v}),\qquad s_v=+1\text{ at a maximum node},\quad s_v=-1\text{ at a minimum node},
\label{sad:eq:edge-weights}
\end{equation}
where $z_v$ collects the children's values. The support $S_v=\{c\in\ch(v):\alpha_{v,c}>0\}$ of $v$ holds the children whose signed values lie strictly beyond the threshold and changes only where a signed child value crosses it. Under \slse{} (\cref{sad:app:prelim}) every weight is positive and $S_v=\ch(v)$.

\subsection{Nonlinear program and its derivatives}
\label{sad:subsec:nlp}
We keep the states as decision variables \citep{bock1984multiple}, $w=(x_1,\ldots,x_T,u_0,\ldots,u_{T-1})\in\R^n$ with $n=T(n_x+n_u)$, and impose the defects $d_t(w)=x_{t+1}-f_h(x_t,u_t)$ as equality constraints. The specification enters through a finite set $\mathcal R$ of rows, one for every conjunct at every time at which the controller imposes it. Each row $r$ is the root node $v_r$ of a conjunct of the evaluation graph at a time $t_r$, and its smoothed value $\widetilde\rho_r(w)$ reads states only. Every temporal operator inside a row is smoothed over its whole window. The constraint vector and the problem are
\begin{equation}
c(w)=\bigl(d_0,\ldots,d_{T-1},(\widetilde\rho_r)_{r\in\mathcal R}\bigr),\qquad
\min_{w}\ E(w)\quad\text{s.t.}\quad d_t(w)=0,\ \ \widetilde\rho_r(w)\ge\delta,
\label{sad:eq:problem}
\end{equation}
with a margin $\delta\ge0$, the initial state $x_0$ given by the measured state, bounds on the states and the controls, and a quadratic cost $E$. An interior-point solver needs the constraint Jacobian $J(w)=Dc(w)$ at every update and, for exact second-order steps, the Hessian of the Lagrangian
\begin{equation}
\mathcal L(w,\lambda,\mu)=\sigma E(w)+\sum_{t=0}^{T-1}\lambda_t^\top d_t(w)+\sum_{r\in\mathcal R}\mu_r\widetilde\rho_r(w),
\label{sad:eq:lagrangian}
\end{equation}
with an objective factor $\sigma$ and multipliers $\lambda_t$ and $\mu_r$.

At every update the solver computes its step from a linear system in the Hessian of $\mathcal L$ and $J$, whose sparsity structure it fixes when a solve starts. With the exact Hessian the step is a Newton step, and a solve from the previous control period's solution ends in a few updates, while a quasi-Newton approximation from successive gradients needs no second derivative and more updates. A solve thus takes the number of updates times the cost of one $J$, one Hessian and one factorization. Automatic differentiation returns products, $u^\top J$, $Jv$ or a Hessian-vector product, and a coloring of a known sparsity pattern lets rows or columns share a product, so the pattern sets their number.

%% file: sections/04-patterns.tex
\section{Sparsity Patterns of the Constraint Jacobian}
\label{sad:sec:patterns}
\subsection{Static and active patterns}
\label{sad:subsec:static-active}
Let $v$ denote an occurrence in the syntax tree of the specification in negation normal form. Its evaluation at sample $t$ is a node $(v,t)$ whose children $\mathcal C(v,t)$ are the nodes of its operands at the samples that its window reads. An until $v_L\,\mathbf U_I\,v_R$ instead has as children one witness node $(v,t,j)$ per $j\in I$, whose children are $(v_R,t+j)$ and $(v_L,t+i)$ for $0\le i\le j$ (\cref{sad:eq:time-children} in \cref{sad:app:patterns}). Conjunctions, always operators and the witness nodes of an until are minimum nodes, and disjunctions, eventually operators and the outer node of an until are maximum nodes. For each predicate $\mu$ we declare a set $A_\mu\subseteq\{1,\ldots,n_x\}$ of state coordinates such that $\partial_j g_\mu(x)=0$ for every $j\notin A_\mu$ on an open domain. A leaf $\ell$ of $\mu$ or $\neg\mu$ at sample $t$ then reads the columns $\mathcal D_\ell=\{(t,j):j\in A_\mu\}$ of $w$, where $(t,j)$ names coordinate $j$ of $x_t$. A leaf at a given state, the measured $x_0$ or a recorded past sample, reads no column, so $\mathcal D_\ell=\emptyset$. Let $S_{v,t}(w)$ be the support of node $(v,t)$ under the edge weights of \cref{sad:eq:edge-weights}, and let $\mathcal A_\ell(w)=\mathcal D_\ell$ at the leaves. The static and the active sets of an internal node are
\begin{align}
\mathcal D_{v,t}&=\bigcup_{c\in\mathcal C(v,t)}\mathcal D_c,\label{sad:eq:static-union}\\
\mathcal A_{v,t}(w)&=\bigcup_{c\in S_{v,t}(w)}\mathcal A_c(w).\label{sad:eq:active-union}
\end{align}

\begin{prop}[Containment of the gradient support]
\label{sad:prop:containment}
For differentiable predicates with the declared coordinate sets and a fixed evaluation graph, $\supp\nabla_w y_{v,t}(w)\subseteq\mathcal A_{v,t}(w)\subseteq\mathcal D_{v,t}$.
\end{prop}
The proof is an induction over the graph with the chain rule (\cref{sad:app:patterns}). For a row $r$ at node $v_r$ and time $t_r$, the indicators $\Pstat_r$ of $\mathcal D_r=\mathcal D_{v_r,t_r}$ and $\Pact_r(w)$ of $\mathcal A_{v_r,t_r}(w)$ are the pattern rows supplied to sparse differentiation. Stacked below the defect pattern they give the static pattern of $J$, which we call the prescribed pattern, and its active pattern at $w$, the supports' pattern.

\subsection{Structural detection and reuse of the active pattern}
\label{sad:subsec:detector}
A structural detector such as CasADi's marks an entry of the Jacobian wherever an output is connected to an input in the expression graph \citep{andersson2019casadi}. It reads the operations of a program and not the values that flow through them. Call a leaf $\ell$ of a row at node $(v,t)$ \emph{attainable} on a set $\mathcal W$ of decision vectors if some $w\in\mathcal W$ has a path from $(v,t)$ to $\ell$ in the evaluation graph along which every edge weight of \cref{sad:eq:edge-weights} is positive.

\begin{prop}[Static pattern and union of the active patterns]
\label{sad:prop:detector}
Fix the evaluation graph of a row at node $(v,t)$, the coordinate sets $A_\mu$ and a set $\mathcal W$ of decision vectors.
\begin{enumerate}
\item With log-sum-exp nodes, whose edge weights are the softmax weights of the node, $\mathcal A_{v,t}(w)=\mathcal D_{v,t}$ for every $w$ in exact arithmetic.
\item With sparsemax nodes, $\bigcup_{w\in\mathcal W}\mathcal A_{v,t}(w)$ is the union of $\mathcal D_\ell$ over the leaves $\ell$ attainable on $\mathcal W$. For a leaf to be attainable it suffices that some $w\in\mathcal W$ and some path from $(v,t)$ to the leaf satisfy the following condition. At every node $u$ of the path other than the leaf, with $m_u$ children and parameter $\gamma_u$, the signed value of the child of $u$ on the path exceeds the largest signed child value of $u$ minus $\gamma_u/m_u$. If every leaf of the row is attainable, the union is $\mathcal D_{v,t}$.
\item Consider a procedure that assigns to the result of every scalar operation of a program the union of the sets assigned to its arguments. Suppose the program computes each leaf $\ell$ from exactly the coordinates in $\mathcal D_\ell$, and the value of every internal node from the values of all of its children and from constants only. Then the procedure assigns $\mathcal D_{v,t}$ to the row, with log-sum-exp nodes and with sparsemax nodes alike. If the program reads further values, the set it assigns contains $\mathcal D_{v,t}$.
\end{enumerate}
\end{prop}
Item 1 holds because every softmax weight is positive. For item 2, unfolding \cref{sad:eq:active-union} to the leaves gives the first claim, and the condition places every child on the path above the threshold of its node, which is at most the largest signed child value minus $\gamma_u/m_u$. Item 3 holds because a union over the arguments of every operation reproduces the recursion of \cref{sad:eq:static-union} whatever the values (details in \cref{sad:app:patterns}). By item 3, a detector that propagates the union of the operands' sets through scalar operations, as CasADi does, returns the static pattern of the Jacobian for sparsemax as it does for log-sum-exp, so exchanging one smoothing for the other inside such a tool does not change that pattern. The zeros of sparsemax depend on the point. A method uses them by reading the supports in one evaluation of the row and building $\Pact(w)$, and it reuses that pattern at the next point once a test confirms that it still holds there. \phantomsection\label{sad:subsec:reuse}Strict gaps $\eta_v$ between the signed child values and the threshold at every node of a row, with Lipschitz bounds $L_v$ of the node inputs in $w$, give the radius $\min_{v:L_v>0}\eta_v/(2L_v)$ in the infinity norm within which every support of the row, and with it the active pattern and a cached coloring, is preserved (\cref{sad:lem:stability,sad:eq:radius}). Without such bounds a runtime test of the cached supports at the new node inputs decides it (\cref{sad:subsec:reuse-details}).

%% file: sections/06-second-order.tex
\section{Sparsity of the Lagrangian Hessian}
\label{sad:sec:second-order}
We derive the pattern of the Hessian of an STL row in the states and its assembly, starting from the Hessian of one smoothed node on a fixed support \citep[Appendix~B.1]{mensch2018differentiable}.

A support region $\mathcal R_{\mathcal S}$ is an open set of trajectories on which every node keeps one support $\mathcal S_v$ with strict inequalities, every signed child value of $S_v$ strictly beyond the threshold and every other one strictly short of it. The statements of this section hold on one such region. Write $J_v$ for the Jacobian of the child values $\zeta_v=z_v$ of node $v$ in the variables, let $k_v=|S_v|$, and let $P_v$ be the sum over the paths from the row to $v$ of the products of their edge weights. On $\mathcal R_{\mathcal S}$ a sparsemax node $v$ is a quadratic polynomial of its children whose Hessian $H_v=\tfrac{s_v}{\gamma_v}\bigl(\Diag(\1_{S_v})-\1_{S_v}\1_{S_v}^\top/k_v\bigr)$ is zero outside $S_v\times S_v$, while the Hessian of a log-sum-exp node is nonzero on all of its children when it has two or more. The Hessian of a row is the sum over the nodes of $P_v\,J_v^\top H_vJ_v$ plus the sum over the leaves of $P_\ell\,\nabla^2z_\ell$, with $z_\ell$ the signed predicate value of leaf $\ell$ (\cref{sad:app:second-order}).

\begin{thm}[Sparsity of a constraint Hessian]
\label{sad:thm:row-support}
Let $L_A=\{\ell:P_\ell>0\}$, the leaves joined to the row through supports only. On $\mathcal R_{\mathcal S}$ the Hessian of a sparsemax row in the leaf values is supported on $L_A\times L_A$, and its Hessian in the states is supported on $\mathcal A_r\times\mathcal A_r$, where $\mathcal A_r=\bigcup_{\ell\in L_A}\mathcal D_\ell$ is the active Jacobian row of \cref{sad:eq:active-union}. The predicate term is block diagonal in time, one block of side $|A_\mu|$ per active leaf. For a log-sum-exp row every weight is positive, $L_A$ is every leaf in the windows, $\mathcal A_r=\mathcal D_r$, and the Hessian is supported on the window clique $\mathcal D_r\times\mathcal D_r$.
\end{thm}

In the sum above a term with $P_v=0$ vanishes, and for $P_v>0$ the matrix $H_v$ keeps only the rows of $J_v$ of the children in $S_v$, whose derivatives reach the leaves of $L_A$ alone. \Cref{sad:app:second-order} has the details, the jump of the Hessian where a support changes (\cref{sad:thm:switch}) and the behavior of Newton's method near a solution on such a surface (\cref{sad:thm:newton-switch}). In the Hessian of the Lagrangian in \cref{sad:eq:lagrangian} the cost couples each control with itself and with its successor and adds a diagonal on the tracked joint positions, each defect term is one dense block on $(x_t,u_t)$ of side $n_x+n_u$, on $u_0$ alone for $t=0$, and each row adds $\mathcal A_r\times\mathcal A_r$ for sparsemax at the current point, or $\mathcal D_r\times\mathcal D_r$ for the static declaration and for log-sum-exp. A star or acyclic coloring needs at least as many colors as the largest clique of the pattern, so the windows of the temporal operators set the bound on the static and log-sum-exp patterns and the supports set it on the active one. The Hessian is assembled by star coloring \citep{gebremedhin2007new} or in closed form from \cref{sad:eq:composition}.

%% file: sections/08-plan.tex
\section{Numerical Experiments}
\label{sad:sec:plan}
\subsection{Experimental setup}
\label{sad:subsec:setup}
\textbf{Comparators.} Every table and figure compares three ways of obtaining the constraint Jacobian and the Hessian of the Lagrangian of the same program. Dense automatic differentiation computes every entry, with $\min\{m,n\}$ products for an $m\times n$ Jacobian and $n$ Hessian-vector products. \slse{} in CasADi with IPOPT is the smooth baseline in its documented form \citep{andersson2019casadi,wachter2006implementation}. CasADi detects the static pattern (\cref{sad:prop:detector}) and colors it, and IPOPT runs with its quasi-Newton approximation of the Hessian (L-BFGS) or with CasADi's exact Hessian. On both scenes below CasADi detects the prescribed Jacobian pattern entry for entry, for the sparsemax program as well (\cref{sad:subsec:app-casadi}), so the supports are what our method adds over this baseline. \smax{} with our patterns, colored and differentiated in JAX~\citep{jax2018github} with asdex \citep{hill2026asdex}, uses the supports' pattern and the exact Hessian of \cref{sad:sec:second-order} with the same IPOPT. The smoothings are matched at equal worst-case error per node (\cref{sad:eq:matching}), and products are counted after a greedy coloring \citep{gebremedhin2007new}.

\textbf{Task.} The plant is a dual-arm manipulator with seven joints per arm, planned on a double integrator in joint space with 14 joint positions, velocities and accelerations per time step, sampled at 0.02~s and bounded. The arms share a hand-off zone divided into cells, balls of radius 0.03~m that only one arm may occupy at a time, so that one arm can follow the other through the zone at a short time gap. The predicate $\mathrm{out}_{a,z}$ states that the 17 distal collision spheres of arm $a$ are outside cell $z$, and $g_{a,p}$ that the hand of arm $a$ is within 0.05~m of waypoint $p$. Both read joint positions only. A schedule assigns every visit $i$ of a cell $z$ to one arm $v$, with $o$ the other arm, and every waypoint $p$ to an arm $a$ with a deadline $d_p$, and the specification has one conjunct per visit and one per waypoint,
\begin{equation}
\mathrm{out}_{o,z}\,\mathbf U_{[0,W_i-1]}\,\mathrm{out}_{v,z}\qquad\text{and}\qquad\mathbf F_{[0,d_p]}\,g_{a,p}.
\label{sad:eq:spec}
\end{equation}
The until conjunct keeps the other arm out of the cell until the visitor has left it, within $W_i$ samples. It enters the program of a control period once its whole window lies inside the horizon (\cref{sad:subsec:app-solver}). Every conjunct is one constraint, its smoothed robustness at or above a margin of 0.02, and the cost penalizes the deviation from a reference motion, the accelerations and their increments.

\textbf{Scenes.} On the \emph{five-cell zone} the arms cross five cells once per loop of 3.2~s, half a loop apart, and $W_i$ is the scheduled stay plus 30 samples, 49 to 86 samples in all. A horizon of 100, 200 and 400 time steps holds 6, 13 and 23 until conjuncts with long windows. On the \emph{64-cell circle} (\cref{sad:fig:circle}) each arm passes 64 cells on a circle once per turn of 3~s, half a turn behind the other, with $W_i=100$. A horizon then holds 28, 121 and 298 until conjuncts with overlapping windows. In the \emph{hold-up} scenario the five-cell zone is timed so that one arm enters a cell 11 to 15 samples after the other has left it. The leading arm is held up by 12 to 20 samples and its schedule moves with it, so the other arm arrives early and the controller has to keep it out of each cell until the leader has left (\cref{sad:fig:replanning}).

\textbf{Protocol.} Every time of the tables and of \cref{sad:fig:replanning}, except the embedded computer's, comes from one workstation, one job at a time on 12 cores, in double precision. A derivative time is the median of up to 30 evaluations after compilation and a repeated solve or closed-loop time the median of its runs (\cref{sad:subsec:app-solver,sad:tab:spread}). Both sides solve the same program up to the smoothing, with the same cost, bounds and IPOPT options, from the solution of the previous control period shifted by one sample or cold from the reference motion. Our solver declares the structure of the supports, tests at every accepted iterate, before its Hessian is evaluated, whether the supports leave it and, if they do, rebuilds on their union and continues from that iterate, all within its reported time. On both sides the reported time of a single solve is that of the solve, with the program of the control period built, and ours compiled, beforehand. In the loops our program is compiled once per horizon, and we report the whole control period against the baseline's solve alone.

%% file: sections/09-results.tex
\subsection{Results}
\label{sad:subsec:results}
\begin{figure}[t]
\centering
\includegraphics{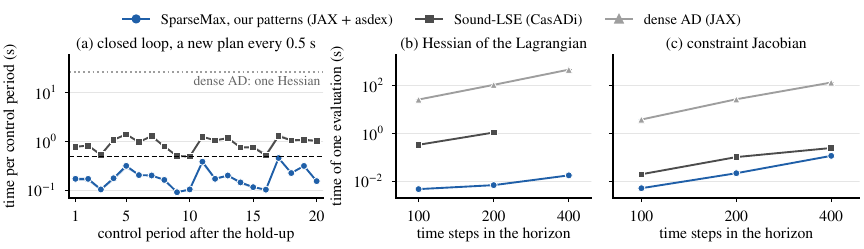}\\[3pt]
\includegraphics[width=0.957\textwidth]{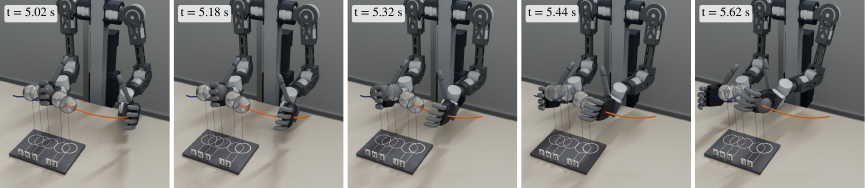}
\caption{(a) Time of every control period of the closed loop on the rigid-body simulation after a hold-up, our whole control period against the baseline's solve alone (medians of six runs; dashed: the control period). (b), (c) Time of one Hessian of the Lagrangian on the five-cell zone and of one constraint Jacobian on the 64-cell circle (medians of up to 30 evaluations). Bottom: the closed-loop motion on the rigid-body simulation, where the right hand (blue path) leaves cell~3 before the left hand (orange path) enters it.}
\label{sad:fig:replanning}
\end{figure}

We demonstrate a model predictive controller that re-plans the joint motion of both arms through the hand-off zone. In the hold-up scenario it starts where the right arm is held up, applies the first 0.5~s of accelerations of each plan to the double integrator and solves again from the shifted plan. Re-planning every 0.5~s over a horizon of 2~s, our whole control period takes 0.11 to 0.40~s in the ten periods after the first solve and lies inside the control period in every one of them. The baseline's solve alone takes 0.46 to 1.35~s. Over a horizon of 4~s the medians are 0.36~s against 6.0~s (\cref{sad:tab:replanning}). In closed loop on a rigid-body simulation of the two arms with their joint drives, re-planning from the simulated state for 10.5~s, every control period after the first takes 0.09 to 0.47~s in six runs against 0.49 to 1.45~s for the baseline's solve alone (\cref{sad:fig:replanning}a). All 33 until constraints and all 21 reach constraints hold in exact semantics on the executed motion, the until constraints with a smallest robustness of 0.28 against $-0.41$ on the motion that ignores the hold-up, and the arms stay at least 38~mm apart (\cref{sad:subsec:app-solver}). At 200 time steps the first solve after the hold-up repairs a plan that violates 6 to 7 until conjuncts, ours 2.5 to 4.4 times sooner than the baseline's (\cref{sad:tab:further-solves}) and at a cost 1.7 to 6.6 per cent lower. The bottom row of \cref{sad:fig:replanning} shows the closed-loop motion at one hand-off of cell~3.

\Cref{sad:tab:solves} gives single solves, with the GMR as the baseline's second smoothing (\cref{sad:subsec:app-gmr}), and its row after a hold-up is the median solve of the ten periods that follow it. From the previous control period's solution the exact Hessian on the supports' structure brings the solve to 3 updates, where IPOPT's quasi-Newton approximation needs 67 to 144. CasADi's exact Hessian gives the baseline the same 3 updates at a higher cost each, so our solve is 6.4 to 27 times shorter than the baseline's shorter one. The supports' structure accounts for a factor of 3.9 to 7.0 of this gain, since the same solver takes that much longer with the prescribed structure declared (\cref{sad:tab:further-solves}). From the reference motion the gain is 1.5 to 8.9, and on the 64-cell circle it is 8.6 from the previous solution with the prescribed structure declared (\cref{sad:subsec:app-solver}). The last column repeats the factor on the CPU of a Jetson AGX Orin, where our solves reproduce to rounding (\cref{sad:subsec:app-orin}).

\begin{table}[t]
\centering
\caption{Single solves on the five-cell zone, as updates of the iterate and seconds, with \slse{} and \gmr{} in CasADi and IPOPT. The factor divides the shorter \slse{} time by ours (star: the baseline's solve ended with an error; dash: not run or ended without a solution).}
\label{sad:tab:solves}
\footnotesize
\setlength{\tabcolsep}{3.5pt}
\begin{tabular}{llcccccc}
\toprule
 & & \multicolumn{2}{c}{\slse} & \gmr & \smax & \multicolumn{2}{c}{Factor}\\
\cmidrule(lr){3-4}\cmidrule(lr){5-5}\cmidrule(lr){6-6}\cmidrule(lr){7-8}
Start & $T$ & Quasi-Newton & Exact Hessian & Quasi-Newton & Ours & Workstation & Orin\\
\midrule
Previous solution & 100 & 67, 0.79 & 3, 1.19 & 75, 0.95 & 3, 0.12 & 6.4 & 4.6\\
 & 200 & 144, 5.53 & 3, 4.13 & 135, 5.16 & 3, 0.22 & 18.4 & 14.8\\
 & 400 & 142, 10.81 & 3, 11.27 & 175, 14.48 & 3, 0.40 & 26.8 & 22.5$^{\ast}$\\
After a 16-sample hold-up & 200 & 112 to 170, 4.09 & -- & -- & 3 to 7, 0.22 & 18.7 & 13.9\\
\bottomrule
\end{tabular}
\end{table}

\begin{figure}[t]
\centering
\includegraphics[scale=0.86]{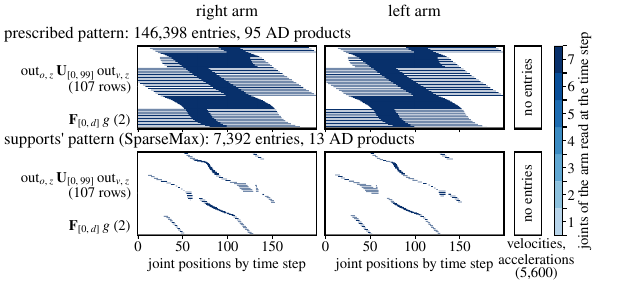}\\[4pt]
\includegraphics{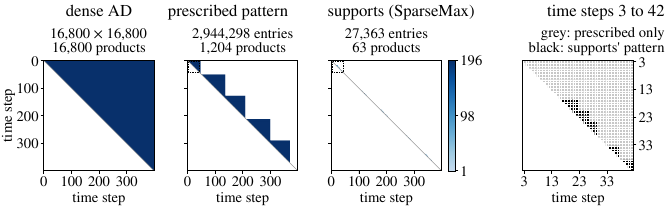}
\caption{Sparsity patterns of the specification's constraints at one control period. Top: the constraint Jacobian on the 64-cell circle with 200 time steps. Bottom: the Hessian of the Lagrangian on the five-cell zone with 400 time steps, on the joint positions, with its Hessian-vector products.}
\label{sad:fig:patterns}
\end{figure}

The derivatives account for these differences. \Cref{sad:fig:patterns} shows what the supports remove from the prescribed pattern, and \cref{sad:tab:counts} counts it in derivative products. On the five-cell zone the prescribed Hessian pattern is a staircase of blocks spanned by the until windows, and the supports leave a narrow band. The Hessian-vector products fall from 1{,}204, the 14 joints times the longest window of 86 samples, to 63, a factor of 19. The Jacobian gains most where many until windows overlap. On the 64-cell circle its products fall from 102 to 13 at 400 time steps, against 130 with CasADi's own coloring, and on the five-cell zone from 8 to 4.

In time, one Hessian of the Lagrangian on the five-cell zone, assembled in closed form on the supports (\cref{sad:sec:second-order}), takes 72 and 160 times less time than CasADi's exact Hessian at 100 and 200 time steps (\cref{sad:fig:replanning}b). On one thread, as CasADi's default evaluation runs, it takes 33 and 73 times less (\cref{sad:tab:threads}). Our Jacobian is 1.9 to 4.7 times shorter than CasADi's with either smoothing on the 64-cell circle and 1.5 to 1.9 times shorter on the five-cell zone at 200 and 400 time steps (\cref{sad:tab:counts}).

%% file: sections/10-conclusion.tex
\section{Conclusion}
\label{sad:sec:conclusion}
Structural detection, as in CasADi, returns the prescribed pattern for sparsemax as for log-sum-exp. The supports lower the Hessian-vector products by a factor of 19 on the five-cell zone and the Jacobian products by a factor of 7.8 on the 64-cell circle. With exact Hessians on the supports' structure a solve from the previous period's solution takes 3 updates and is 6.4 to 27 times shorter than log-sum-exp in CasADi with IPOPT. Re-planning every 0.5~s over 2~s after a hold-up, our whole control period takes at most 0.40~s against up to 1.35~s for the baseline's solve alone, and every constraint of the specification holds in closed loop on a rigid-body simulation of the arms.

%% file: sections/A-appendix.tex
\section{Extended Related Work}
\label{sad:app:related}
STL robustness reduces Boolean and temporal operators to nested extrema over predicate values \citep{maler2004monitoring,donze2010robust}, and differentiable computation graphs expose it to gradient-based methods \citep{leung2023backpropagation,kapoor2025stlcg++}. The log-sum-exp robustness replaces each extremum by a soft one \citep{pant2017smooth}. The softmax-mean measure pairs a log-sum-exp minimum with a softmax-weighted mean as the maximum, both under-approximations, so a positive value implies satisfaction for every parameter value \citep{gilpin2020smooth}. A comparison of four smooth measures composes per-node error bounds through the specification and allows different parameters at different levels of it \citep{welikala2023smooth}. Average-based measures form a further family \citep{mehdipour2019arithmetic,mehdipour2025generalized,uzun2024optimization}. The sparsemax lower extrema of our earlier paper \citep{damera2026sparsemax} are built from the quadratically regularized maximum \citep{mensch2018differentiable}, whose gradient is the Euclidean projection onto the simplex \citep{martins2016Sparsemax}. That paper proves that the resulting robustness is a lower bound of the exact one with an error known before optimization, and that its weight on violating samples a gap beyond the others is bounded below independently of the horizon. This paper builds on both results and shows what their supports give for the cost of the derivatives of a constrained planner.

An exact reformulation of the extrema avoids the approximation error of a smooth measure \citep{han2025exact}. A constraint on a minimum becomes one constraint per operand, and a constraint on a maximum becomes a constraint on a convex combination of the operands whose weights are decision variables. Applied at every nested extremum with auxiliary variables, the two rules give a smooth nonlinear program whose feasible trajectories are those of the original constraint. The number of added variables and constraints grows linearly with the size of the tree of nested extrema. We keep each constraint as one smoothed robustness function of the trajectory, which adds no decision variable and carries the error bound of that paper. We derive the sparsity patterns of its derivatives from the specification and its supports.

Sparse automatic differentiation recovers a Jacobian or a Hessian from one product per color of a known sparsity pattern \citep{curtis1974estimation,gebremedhin2005color,gebremedhin2007new}. The pattern is declared by the user or detected from the program \citep{hill2025sparser,hill2026asdex}. Structural detection, also called global, propagates dependencies through the operations and returns a pattern that is valid at every input, while local detection also evaluates the operations at one input and returns the pattern at that input \citep{hill2025sparser}. A local pattern can be reused at another input when it contains the pattern there, and the margin of \cref{sad:subsec:reuse} supplies such a test for the supports' pattern, which we read from the forward evaluation of the constraints without tracing the program. CasADi detects the pattern of a Jacobian by propagating dependencies through its expression graph and treats a Hessian as the Jacobian of a gradient with symmetry exploited \citep{andersson2019casadi,wachter2006implementation}. We therefore count it as the established baseline with a sparse pattern and not as a dense one. Direct multiple shooting enters the dynamics through local defect constraints \citep{bock1984multiple}, and STL specifications have been enforced in receding-horizon control \citep{sadraddini2015robust,raman2014model}.

\section{STL Semantics and the Smoothed Operators}
\label{sad:app:prelim}
The model has a fixed initial state $x_0$. The controls $U=(u_0,\ldots,u_{T-1})$ lie in a box $\mathcal U$ of actuator bounds and generate the sampled trajectory $X=(x_0,\ldots,x_T)$ over the horizon $H=Th$. For the grammar of \cref{sad:eq:grammar}, the robustness at sample $t$ is defined recursively from the predicate values by
\begin{equation}
\begin{aligned}
\rho^{\phi\wedge\psi}_t&=\min\{\rho^\phi_t,\rho^\psi_t\},\qquad
\rho^{\phi\vee\psi}_t=\max\{\rho^\phi_t,\rho^\psi_t\},\\
\rho^{\mathbf G_I\phi}_t&=\min_{j\in I}\rho^\phi_{t+j},\qquad
\rho^{\mathbf F_I\phi}_t=\max_{j\in I}\rho^\phi_{t+j},\\
\rho^{\phi\,\mathbf U_I\,\psi}_t&=\max_{j\in I}\,\min\Bigl\{\rho^\psi_{t+j},\ \min_{0\le i\le j}\rho^\phi_{t+i}\Bigr\}.
\end{aligned}
\label{sad:eq:semantics-full}
\end{equation}

Negation normal form pushes every negation to a predicate through De Morgan's laws, the duality of always with eventually and, for a negated until, the identity $\neg(\phi\,\mathbf U_I\,\psi)\equiv\bigwedge_{j\in I}\bigl(\mathbf F_{\{j\}}\neg\psi\vee\mathbf F_{\{0,\ldots,j\}}\neg\phi\bigr)$, which holds in the sampled exact semantics. The conversion is made in exact semantics before any smoothing, because the negation of a lower bound of a robustness is an upper bound of the negated robustness.

Unrolling the recursion of \cref{sad:eq:semantics-full} over the samples gives a finite acyclic evaluation graph whose leaves are predicate values and whose internal nodes are minima and maxima. A node is evaluated only where every sample its window reads exists, so every window keeps its full number of children and the horizon of a specification is at most $H$. We write $\ch(v)$ for the children of a node $v$ and $m_v$ for their number. An until node at sample $t$ is a maximum over its witnesses $j\in I$, and the child of witness $j$ is a minimum over the $j+2$ values $\rho^\psi_{t+j},\rho^\phi_t,\ldots,\rho^\phi_{t+j}$. Its inner minima therefore differ in their numbers of children, which an implementation handles with masked windows.

Each node is evaluated as one operator over all of its children, because the smoothed operators are not associative. At a node whose $k$ extreme children lie a gap $\Delta$ beyond its $N$ other children, the sparsemax weight on the $k$ children is at least $\min\{1,k\Delta/\gamma\}$ for every $N$. At equal worst-case error the weight of the sound log-sum-exp on them tends to zero as $N$ grows when the other children lie within that error of them \citep{damera2026sparsemax}.

We write $\rho_v$ for the exact value of a node and $\widetilde\rho_v$ for its smoothed value. The budget of a node is $B_v=b_{m_v}(\gamma_v)+\max_{u\in\ch(v)}B_u$, with $B_v=0$ at an exact predicate leaf.

\begin{thm}[Lower bound, \citealp{damera2026sparsemax}]
\label{sad:thm:lower}
On a finite acyclic evaluation graph of minima and maxima with exact leaves, every node satisfies $0\le\rho_v-\widetilde\rho_v\le B_v$, and at the root $B_v$ is the largest sum of node bands along a root-to-leaf path.
\end{thm}

Our baseline (\slse) is the sound log-sum-exp. Its maximum $L_\beta(z)=\beta^{-1}\log\sum_ie^{\beta z_i}-\log m/\beta$ lies below the maximum by at most $\log m/\beta$ \citep{mehdipour2025generalized,damera2026sparsemax}, as does the log-sum-exp minimum $-\beta^{-1}\log\sum_ie^{-\beta z_i}$ below the minimum \citep{gilpin2020smooth}. Its gradient is the softmax of $\beta z$, which is positive at every entry, so every child of a log-sum-exp node is in its support. The two families are compared at equal worst-case error $\varepsilon$ per node with $m\ge2$ children,
\begin{equation}
\beta=\frac{\log m}{\varepsilon},\qquad
\gamma=\frac{2\varepsilon}{1-1/m}.
\label{sad:eq:matching}
\end{equation}

Multiple shooting places every predicate leaf at sample $t$ on the variable $x_t$ alone, so the gradient of a row touches only the states its windows read. Under single shooting a late predicate depends on every earlier control, and the adjoint of even a sparse forcing propagates through every earlier step of the dynamics. A solver point that meets the constraints of \cref{sad:eq:problem} has rows of at least $\delta$ on its own states, which by \cref{sad:thm:lower} gives exact rows of at least $\delta$ on those states.

The defect $d_0$ has no block in the given $x_0$. In the rows of the defect $d_t$ with $t\ge1$ the only nonzero blocks are
\begin{equation}
\left[
\frac{\partial d_t}{\partial x_t}\ \
\frac{\partial d_t}{\partial x_{t+1}}\ \
\frac{\partial d_t}{\partial u_t}
\right]
=\bigl[-D_x f_h\ \ I\ \ -D_u f_h\bigr].
\label{sad:eq:defect-blocks}
\end{equation}

Without using any internal structure of the dynamics, two palettes of $n_x$ colors alternating over even and odd samples and $n_u$ further colors for the controls give a valid column coloring of the defect rows. It has $2n_x+n_u$ colors, a number independent of the horizon.

\section{Jacobian Patterns: Proofs, a Worked Example and Reuse}
\label{sad:app:patterns}
The children of the node $(v,t)$ of an occurrence $v$ in the syntax tree at sample $t$ are
\begin{equation}
\mathcal C(v,t)=
\begin{cases}
\{(v_L,t),(v_R,t)\},&v\text{ is }\wedge\text{ or }\vee,\\
\{(v_C,t+j):j\in I\},&v\text{ is }\mathbf F_I\text{ or }\mathbf G_I,\\
\{(v,t,j):j\in I\},&v\text{ is }v_L\,\mathbf U_I\,v_R,\\
\{(v_R,t+j)\}\cup\{(v_L,t+i):0\le i\le j\},&\text{the witness node }(v,t,j)\text{ of an until},
\end{cases}
\label{sad:eq:time-children}
\end{equation}
where $v_L$, $v_R$ and $v_C$ are the left, right and only children of $v$ in the syntax tree.

\begin{proof}[Proof of \cref{sad:prop:containment}]
At a leaf the claim is the coordinate declaration. At an internal node the chain rule gives $\nabla_w y_{v,t}=\sum_{c\in\mathcal C(v,t)}\alpha_{v,t,c}\nabla_w y_c$, and edges of weight zero contribute nothing. By induction every surviving child gradient lies in the union of \cref{sad:eq:active-union}, which is a subset of the union of \cref{sad:eq:static-union}.
\end{proof}

\begin{proof}[Proof of \cref{sad:prop:detector}]
Every log-sum-exp weight is positive, so every support is the whole child set and \cref{sad:eq:active-union} is \cref{sad:eq:static-union}, which gives item 1. For item 2, unfold \cref{sad:eq:active-union} down to the leaves. The set $\mathcal A_{v,t}(w)$ is the union of $\mathcal D_\ell$ over the leaves reached from $(v,t)$ through support children only, and these are the leaves joined to $(v,t)$ by a path of positive weights at $w$. The union over $\mathcal W$ is the first claim. For the sufficient condition, let $z\in\R^m$ hold the signed child values of a node, $\theta$ its threshold and $k\ge1$ the size of its support. In \cref{sad:eq:threshold} exactly $k$ terms are positive and each is at most $z_{\max}-\theta$, so $\gamma\le k(z_{\max}-\theta)$ and $\theta\le z_{\max}-\gamma/k\le z_{\max}-\gamma/m$. A child whose signed value exceeds $z_{\max}-\gamma/m$ therefore lies above the threshold and has a positive weight, and under the condition this holds at every edge of the path. Unfolding \cref{sad:eq:static-union} in the same way shows that $\mathcal D_{v,t}$ is the union of $\mathcal D_\ell$ over all leaves of the row, which gives the last claim. For item 3, the result of a scalar operation receives the union of the sets of its arguments. The value of a node is computed from the values of all of its children and from constants, so it receives the union of the sets of its children, which is the recursion of \cref{sad:eq:static-union} from the leaf sets $\mathcal D_\ell$. A further value that the program reads can only add to a set. Every program that evaluates a sparsemax node correctly for all child values reads every child, because by \cref{sad:eq:threshold} each child enters the support, and with it the value, for some child values.
\end{proof}

The condition of item 2 of \cref{sad:prop:detector} holds at any $w$ at which every node on the path has the largest signed smoothed value among the children of its parent, ties included.

\subsection{Worked example}
\label{sad:subsec:worked}
\input{figures/nnf-pattern-tree}
Negation normal form turns the specification on the left of
\begin{equation}
\neg\bigl(\mathbf G_{[1,2]}\neg\mu\vee\mathbf F_{[0,1]}\nu\bigr)\equiv\varphi,\qquad
\varphi=\mathbf F_{[1,2]}\mu\wedge\mathbf G_{[0,1]}\neg\nu
\label{sad:eq:nnf-example}
\end{equation}
into $\varphi$. We evaluate $\varphi$ at $t=1$ with states $x_t=(a_t,b_t,c_t)$ and the affine predicates $g_\mu(x_t)=a_t+b_t$ and $g_\nu(x_t)=b_t+c_t$. The temporal nodes read the leaf pairs $(g_\mu(x_2),g_\mu(x_3))$ and $(-g_\nu(x_1),-g_\nu(x_2))$, and \cref{sad:fig:parse-tree} shows the graph. With $\gamma=1$ at every node, the three internal values are
\begin{equation}
y_F=Q_1^+\bigl(g_\mu(x_2),g_\mu(x_3)\bigr),\qquad
y_G=Q_1^-\bigl(-g_\nu(x_1),-g_\nu(x_2)\bigr),\qquad
y_\varphi=Q_1^-(y_F,y_G).
\label{sad:eq:example-nodes}
\end{equation}
At the states
\begin{equation}
x_1=(0,-1,0),\qquad x_2=(1,1,-4),\qquad x_3=(1/2,0,0)
\label{sad:eq:example-states}
\end{equation}
the leaf scores are $(2,1/2,1,3)$. The gaps inside the two temporal nodes, $3/2$ and $2$, exceed $\gamma$, and at the root the projection of $(-7/4,-1)$ has the threshold $-15/8$, so both branches of the root stay active,
\begin{equation}
\alpha_F=(1,0),\ \ y_F=7/4,\qquad
\alpha_G=(1,0),\ \ y_G=1,\qquad
\alpha_\varphi=(1/8,7/8),\ \ y_\varphi=63/64.
\label{sad:eq:example-weights}
\end{equation}
The exact robustness is $1$, consistent with the lower value $63/64$.

Restrict the columns to $\bar w=(a_1,b_1,c_1,a_2,b_2,c_2,a_3,b_3,c_3)$, outside which every column of the Jacobian is zero, and stack the intermediate outputs $(y_F,y_G,y_\varphi)$ to display the construction. The static and active patterns are
\begin{equation}
\Pstat=
\begin{bmatrix}
0&0&0&1&1&0&1&1&0\\
0&1&1&0&1&1&0&0&0\\
0&1&1&1&1&1&1&1&0
\end{bmatrix},\qquad
\Pact=
\begin{bmatrix}
0&0&0&1&1&0&0&0&0\\
0&1&1&0&0&0&0&0&0\\
0&1&1&1&1&0&0&0&0
\end{bmatrix}.
\label{sad:eq:example-patterns}
\end{equation}
The static root row has seven columns and the active root row four, because the edges to $g_\mu(x_3)$ and $-g_\nu(x_2)$ have weight zero. Their removal drops $a_3$, $b_3$ and $c_2$ but keeps $b_2$, which the other predicate reads at sample $2$. Pruning one occurrence of a predicate must therefore never erase a column that another active path uses. The chain rule gives the Jacobian $\bar J=\partial(y_F,y_G,y_\varphi)/\partial\bar w$,
\begin{equation}
\bar J=\begin{bmatrix}
0&0&0&1&1&0&0&0&0\\
0&-1&-1&0&0&0&0&0&0\\
0&-\frac78&-\frac78&\frac18&\frac18&0&0&0&0
\end{bmatrix}.
\label{sad:eq:example-jacobian}
\end{equation}
At this trajectory the exact root selects only the always branch, whereas the smoothed root keeps both, so the active pattern describes the derivative of the smoothed robustness.

\subsection{Recovery from compressed products}
\label{sad:subsec:reconstruction}
\begin{prop}[Recovery from compressed products]
\label{sad:prop:reconstruction}
Suppose $P$ contains the support of $J$. Assign column colors $\kappa(j)\in\{1,\ldots,q\}$ so that no two columns that share a marked row receive the same color, set $V_{j,a}=\1\{\kappa(j)=a\}$ and compute $Y=JV$. Then $J_{ij}=Y_{i,\kappa(j)}$ wherever $P_{ij}=1$, and every other entry of $J$ is zero.
\end{prop}
\begin{proof}
The entry $Y_{i,\kappa(j)}$ sums $J_{i\ell}$ over the columns $\ell$ of color $\kappa(j)$. Validity of the pattern removes the unmarked columns, and the coloring condition leaves only $\ell=j$.
\end{proof}

By \cref{sad:prop:reconstruction}, $q$ directional derivatives reconstruct $J$, and the transposed argument with row colors uses $q$ vector-Jacobian products. In the active pattern of \cref{sad:eq:example-patterns} the first two rows have disjoint columns and can share one reverse product. With the output seeds $R=\bigl[\begin{smallmatrix}1&1&0\\0&0&1\end{smallmatrix}\bigr]$ and $B=R\bar J$, the first compressed row is $\bar J_F+\bar J_G$, its columns $a_2,b_2$ recover $\bar J_F$ and its columns $b_1,c_1$ recover $\bar J_G$, and the second compressed row is the root row. In the static pattern the first two rows overlap at $b_2$ and cannot share a product. An omitted nonzero corrupts another recovered entry through a color collision, so the reuse of a pattern at another point requires the test of \cref{sad:subsec:reuse-details}.

\subsection{Reuse of an active pattern}
\label{sad:subsec:reuse-details}
A change of the child values can bring into the support a child that the current support discards. A strict gap between each child value and the threshold bounds how far the values may move before that happens.

\begin{lemma}[Stability of the support]
\label{sad:lem:stability}
Let a node have signed child values $z$, threshold $\theta$, support $S=\{i:z_i>\theta\}$ and margin $\eta=\min_i|z_i-\theta|>0$. For a fixed $\gamma$ and a fixed number of children, every perturbation with $\|\delta z\|_\infty<\eta/2$ preserves $S$.
\end{lemma}
\begin{proof}[Proof of \cref{sad:lem:stability}]
The threshold candidate that keeps the old support is $\theta'=\theta+|S|^{-1}\sum_{i\in S}\delta z_i$, whose displacement is at most $\|\delta z\|_\infty$. Each child value minus threshold therefore changes by less than $\eta$ and keeps its strict sign. The candidate satisfies the normalization in \cref{sad:eq:threshold} on $S$, so the uniqueness of the projection makes it the new threshold.
\end{proof}

The test covers every node of a row, because the inputs of its descendants change too. Assume strict margins $\eta_v>0$ at every node, and suppose that every node input map obeys $\|z_v(w+\delta w)-z_v(w)\|_\infty\le L_v\|\delta w\|_\infty$ on a stated neighbourhood. Applying \cref{sad:lem:stability} at every node keeps all supports for
\begin{equation}
\|\delta w\|_\infty<\min_{v:L_v>0}\frac{\eta_v}{2L_v},
\label{sad:eq:radius}
\end{equation}
as long as the step stays in the neighbourhood. The bounds $L_v$ follow from Lipschitz bounds of the leaves and the nonexpansiveness of the lower extrema in the infinity norm.

In the worked example the margins are $\eta_F=1/2$, $\eta_G=1$ and $\eta_\varphi=1/8$, each affine leaf is $2$-Lipschitz in the infinity norm, and the lower extrema keep that bound for every node input, so $\|\delta\bar w\|_\infty<1/32$ keeps the active pattern of \cref{sad:eq:example-patterns}.

Where no Lipschitz bounds are available, the runtime check evaluates every node input, computes the thresholds that the cached supports imply and tests every active and inactive inequality. A failed test triggers a new support detection and pattern construction before any compressed product.

We checked these constructions on random specifications with every operator, until included. No entry of a dense Jacobian lay outside the static pattern, nor outside the active pattern where the supports were strict, and recovery through every coloring matched dense differentiation to rounding in double and single precision. Of 1120 random perturbations larger than the reuse radius, 321 changed a support, and the runtime check rejected every one of them. Forcing the cached coloring at those 321 points gave a wrong Jacobian at 154 of them in double precision, with relative errors up to 0.72.

\section{Hessian: Proofs, Support Changes and Newton's Method}
\label{sad:app:second-order}

\begin{lemma}[Hessian of a sparsemax node]
\label{sad:lem:node-hessian}
On $\mathcal R_{\mathcal S}$ the sparsemax node $v$ is a quadratic polynomial of its children with Hessian
\begin{equation}
H_v=\frac{s_v}{\gamma_v}\Bigl(\Diag(\1_{S_v})-\frac{\1_{S_v}\1_{S_v}^\top}{k_v}\Bigr),
\label{sad:eq:node-hessian}
\end{equation}
which is zero outside $S_v\times S_v$ and zero when $k_v=1$. A log-sum-exp node at equal error has $H_v=s_v\beta_v(\Diag\pi-\pi\pi^\top)$ with $\pi$ the softmax of $\pm\beta_v\zeta_v$, which is nonzero on all of its children for $m_v\ge2$.
\end{lemma}

\begin{prop}[Hessian of a constraint]
\label{sad:prop:composition}
On $\mathcal R_{\mathcal S}$ the Hessian of a row $r$ is
\begin{equation}
\nabla^2\widetilde\rho_r=\sum_{v}P_v\,J_v^\top H_vJ_v+\sum_{\ell}P_\ell\,\nabla^2z_\ell,
\label{sad:eq:composition}
\end{equation}
where the second sum runs over the leaves and $z_\ell$ is the signed predicate value of leaf $\ell$.
\end{prop}

We use the published Hessian of one smoothed node on a fixed support. For the quadratically regularized maximum it is $(\Diag(s)-ss^\top/\|s\|_1)/\gamma$ with $s$ the indicator of the support \citep[Appendix~B.1]{mensch2018differentiable}, the Jacobian of sparsemax has the same bracket \citep[Eq.~12]{martins2016Sparsemax}, and Hessian-vector products through a graph of nested smoothed maxima cost as much as its gradient \citep[Section~3.4]{mensch2018differentiable}.

Star coloring recovers the entries directly from the compressed products \citep{gebremedhin2007new}. Each compressed product is one forward-over-reverse pass through the Lagrangian, whose cost grows linearly with $T$. The closed-form assembly instead evaluates \cref{sad:eq:composition} from the supports and from the gradients and Hessians of the active leaves alone. On the five-cell zone at 100 to 400 time steps, one Hessian of the Lagrangian by star coloring takes 12 to 15 times less time for sparsemax on the supports' pattern than for log-sum-exp on the prescribed pattern, both in our implementation. The closed-form assembly on the supports is a further 130 to 175 times shorter.

\begin{proof}[Proof of \cref{sad:prop:composition}]
Every map is twice differentiable on $\mathcal R_{\mathcal S}$. At the row, $\nabla^2y_r=J_r^\top H_rJ_r+\sum_c\alpha_{r,c}\nabla^2y_c$. Applying the same identity at every node in topological order multiplies the term of node $v$ by the sum over the paths to $v$ of the products of the edge weights, which is $P_v$, and a leaf contributes $P_\ell\nabla^2z_\ell$. A node reached by several paths collects each path once.
\end{proof}

\begin{proof}[Proof of \cref{sad:thm:row-support}]
In the sum of \cref{sad:prop:composition}, terms with $P_v=0$ vanish. For $P_v>0$, \cref{sad:lem:node-hessian} keeps only the rows of $J_v$ that belong to children $c\in S_v$, each of which has $P_c\ge P_v\alpha_{v,c}>0$. The derivative $\partial y_c/\partial z_\ell$ is a sum over support paths from $c$ to $\ell$, so it vanishes unless $\ell\in L_A$. The state form follows from $z_\ell=\pm g_\mu(x_t)$, whose gradient lies in $\mathcal D_\ell$.
\end{proof}

For an always over the whole horizon on $|A_\mu|$ coordinates, the log-sum-exp block has side $|A_\mu|T$, while the sparsemax block has side $|A_\mu|$ times the number of samples in the support.

Let the evaluation graph below the row be a tree with $|L_A|\ge2$, and let $v$ be the first node with $k_v\ge2$ reached from the row. In the leaf values, two leaves $a,b\in L_A$ below different children $e\ne e'$ of $v$ receive the entry $-s_vP_v\,\partial_{z_a}\zeta_e\,\partial_{z_b}\zeta_{e'}/(\gamma_vk_v)$ and no other term, which is nonzero on all of $\mathcal R_{\mathcal S}$ because its factors are products of positive weights. Every other entry of $L_A\times L_A$ collects terms of several nodes whose signs can differ, and it vanishes only on a closed set of leaf values of measure zero.

\begin{thm}[Hessian jump at a support change]
\label{sad:thm:switch}
Let $X^\ast$ lie on the common boundary of two support regions that differ at one node $v$, with $\mathcal S'_v=\mathcal S_v\cup\{j\}$ and $k=|\mathcal S_v|$. The row is continuously differentiable at $X^\ast$, the one-sided Hessians $H_{\mathcal S}$ and $H_{\mathcal S'}$ exist as the limits of the Hessian from inside each region, and
\begin{equation}
H_{\mathcal S'}-H_{\mathcal S}=\frac{s_vP_v}{\gamma_v\,k(k+1)}\,qq^\top,\qquad
q=k\,\nabla\zeta_{v,j}-\sum_{i\in\mathcal S_v}\nabla\zeta_{v,i},
\label{sad:eq:switch}
\end{equation}
a jump of rank at most one along the normal of the switching surface.
\end{thm}
\begin{proof}
Freezing every support at $\mathcal S$ gives a function that is twice differentiable near $X^\ast$ and equals the row on $\mathcal R_{\mathcal S}$. At $X^\ast$ the weight of $j$ is zero on both sides, so the weights, the path weights and the child gradients of the two frozen functions agree, and in \cref{sad:eq:composition} only $H_v$ differs. A direct computation gives $H(\mathcal S_v\cup\{j\})-H(\mathcal S_v)=\tfrac{s_v}{\gamma_v}uu^\top$ with $u=(k\,e_j-\1_{\mathcal S_v})/\sqrt{k(k+1)}$, and $J_v^\top uu^\top J_v=qq^\top/(k(k+1))$.
\end{proof}

Simultaneous switches at different nodes add their terms, and a switch at a node with $P_v=0$ changes nothing.

\subsection{Local convergence of Newton's method at a support change}
\label{sad:subsec:newton-switch}
A solution can lie on the boundary between support regions, where the Hessian of a row has two or more one-sided values. Newton's method does not need the Hessian to exist there. At each point the gradient of every row coincides with one of finitely many continuously differentiable maps, one per choice of supports, and a Newton step with the Hessian of the current point's own supports is a Newton step on one of these maps. Every map that can be chosen near the solution vanishes at the solution, and where the Jacobian of each of them is nonsingular there, each such step contracts toward it.

Write \cref{sad:eq:problem} with nonnegative slack variables for the rows and the bounds. The variables $\xi\in\R^N$ then collect $w$ and the slacks, the constraints are equalities $G(\xi)=0$, and $\xi_i\ge0$ for $i$ in an index set $\mathcal B$. For a barrier parameter $\omega>0$ an interior-point method minimizes $E(\xi)-\omega\sum_{i\in\mathcal B}\ln\xi_i$ subject to $G(\xi)=0$. Let $\lambda$ be the multipliers of $G$ and $\chi_{\mathcal B}=(\chi_i)_{i\in\mathcal B}$ those of the bounds, and let $\chi\in\R^N$ extend $\chi_{\mathcal B}$ by zeros. The optimality conditions of this barrier problem are the primal-dual equations
\begin{equation}
F_\omega(\xi,\lambda,\chi_{\mathcal B})=
\begin{pmatrix}
\nabla E(\xi)+DG(\xi)^\top\lambda-\chi\\
G(\xi)\\
(\xi_i\chi_i-\omega)_{i\in\mathcal B}
\end{pmatrix}
=0,\qquad \xi_{\mathcal B}>0,\ \ \chi_{\mathcal B}>0.
\label{sad:eq:primal-dual}
\end{equation}
We write $\Xi=(\xi,\lambda,\chi_{\mathcal B})$.

For a set $S$ of $k$ entries of $z\in\R^m$ let $\theta_S(z)=(\sum_{i\in S}z_i-\gamma)/k$ and $q_S(z)=\theta_S(z)+\tfrac1{2\gamma}\sum_{i\in S}(z_i-\theta_S(z))^2$, a quadratic polynomial of the entries in $S$. By the definition of $M_\gamma$ in \cref{sad:subsec:lower} and by \cref{sad:eq:threshold}, $M_\gamma(z)=q_S(z)$ wherever the support of $z$ is $S$. An assignment $\mathcal S$ gives every internal node $v$ of every row a nonempty set $\mathcal S_v$ of its children. The frozen row $\widetilde\rho^{\mathcal S}_r$ evaluates every node by \cref{sad:eq:lower} with $q_{\mathcal S_v}$ in place of $M_{\gamma_v}$, and $G^{\mathcal S}$ and $F^{\mathcal S}_\omega$ are $G$ and $F_\omega$ with every row frozen at $\mathcal S$. We call $\mathcal S$ active at $\xi$ when $G^{\mathcal S}(\xi)=G(\xi)$ and $DG^{\mathcal S}(\xi)=DG(\xi)$.

\begin{lemma}[Constraints with fixed supports]
\label{sad:lem:frozen}
Let the predicates and the model $f_h$ be twice continuously differentiable. There are finitely many assignments, every $F^{\mathcal S}_\omega$ is continuously differentiable, and the assignment $\mathcal S(\xi)$ of the supports at $\xi$ is active at $\xi$ for every $\xi$. An assignment that is not active at $\bar\xi$ is not active at any point near $\bar\xi$.
\end{lemma}
\begin{proof}
A frozen row composes quadratic polynomials with twice continuously differentiable predicates, and the defects are twice continuously differentiable with the model, so $G^{\mathcal S}$ is twice and $F^{\mathcal S}_\omega$ once continuously differentiable. At a point whose support at a node is $S$, the operator of that node equals its frozen form. The identity $\sum_{i\in S}(z_i-\theta_S(z))=\gamma$ gives $\partial q_S/\partial z_i=(z_i-\theta_S(z))/\gamma$ for $i\in S$ and zero otherwise, which are the edge weights of \cref{sad:eq:edge-weights} when $S$ is the support. Induction from the leaves shows that every row and its frozen row have the same value and the same gradient at the point's own supports. The argument covers a point at which a child value equals the threshold, because the support is defined by a strict inequality. The last statement holds because $G$, $DG$, $G^{\mathcal S}$ and $DG^{\mathcal S}$ are continuous.
\end{proof}

The Jacobian $DF^{\mathcal S}_\omega$ holds the Hessian of the frozen Lagrangian, $W^{\mathcal S}(\xi,\lambda)=\nabla^2E(\xi)+\sum_j\lambda_j\nabla^2G^{\mathcal S}_j(\xi)$, in its first block. At $\mathcal S=\mathcal S(\xi)$ and a point $\xi$ at which no child value equals its node's threshold, this block is the Hessian that automatic differentiation returns at $\xi$ in exact arithmetic. Newton's method for a system of piecewise continuously differentiable equations, with the Jacobian of a piece that contains the current point, converges quadratically from every start near a solution if the Jacobians of the pieces that contain the solution are nonsingular there and Lipschitz continuous near it \citep{kojima1986extension}. \Cref{sad:thm:newton-switch} applies this argument to the barrier equations, whose pieces are the equations with fixed supports, and adds the superlinear rate for Jacobians that are only continuous and the positivity of the slacks and the multipliers.

\begin{thm}[Local convergence of Newton's method]
\label{sad:thm:newton-switch}
Let the predicates and the model $f_h$ be twice continuously differentiable, and let $\bar\Xi=(\bar\xi,\bar\lambda,\bar\chi_{\mathcal B})$ solve \cref{sad:eq:primal-dual}. Let $\mathcal N(\bar\xi)$ be the set of assignments $\mathcal S$ for which every neighbourhood of $\bar\xi$ holds a point $\xi$ with $\mathcal S(\xi)=\mathcal S$, and suppose that $DF^{\mathcal S}_\omega(\bar\Xi)$ is nonsingular for every $\mathcal S\in\mathcal N(\bar\xi)$. Then there is a ball $U$ around $\bar\Xi$ such that for every $\Xi\in U$ the step with the supports of the current point,
\begin{equation}
\Xi^+=\Xi-DF^{\mathcal S(\xi)}_\omega(\Xi)^{-1}F_\omega(\Xi),
\label{sad:eq:newton-step}
\end{equation}
is defined, lies in $U$, keeps $\xi^+_{\mathcal B}$ and $\chi^+_{\mathcal B}$ positive and satisfies $\|\Xi^+-\bar\Xi\|\le\tfrac12\|\Xi-\bar\Xi\|$, with $\|\Xi^+-\bar\Xi\|/\|\Xi-\bar\Xi\|\to0$ as $\Xi\to\bar\Xi$. The iteration that uses the Hessian of each point's own supports therefore converges to $\bar\Xi$ superlinearly from every start in $U$, for any number of children at their thresholds at $\bar\xi$ and however often the iterates change support. The convergence is quadratic when the predicates and the model are three times continuously differentiable.
\end{thm}
\begin{proof}
There are finitely many assignments, so there is a ball $U_0$ around $\bar\Xi$ on which the supports $\mathcal S(\xi)$ of every point belong to $\mathcal N(\bar\xi)$. Every member of $\mathcal N(\bar\xi)$ is active at $\bar\xi$. By \cref{sad:lem:frozen} it is active at points that converge to $\bar\xi$, and an assignment that is not active at $\bar\xi$ is not active near it. The Jacobians of the members are continuous and nonsingular at $\bar\Xi$, so $U_0$ can be chosen such that $\|DF^{\mathcal S}_\omega(\Xi)^{-1}\|\le M$ on it for every $\mathcal S\in\mathcal N(\bar\xi)$. Take $\Xi\in U_0$ and $\mathcal S=\mathcal S(\xi)$. Then $F_\omega(\Xi)=F^{\mathcal S}_\omega(\Xi)$ because $\mathcal S$ is active at $\xi$, and $F^{\mathcal S}_\omega(\bar\Xi)=F_\omega(\bar\Xi)=0$ because $\mathcal S$ is active at $\bar\xi$. The step \cref{sad:eq:newton-step} therefore satisfies
\begin{equation}
\Xi^+-\bar\Xi=DF^{\mathcal S}_\omega(\Xi)^{-1}\int_0^1\Bigl[DF^{\mathcal S}_\omega(\Xi)-DF^{\mathcal S}_\omega\bigl(\bar\Xi+t(\Xi-\bar\Xi)\bigr)\Bigr]dt\,(\Xi-\bar\Xi).
\label{sad:eq:newton-error}
\end{equation}
The norm of the bracket in \cref{sad:eq:newton-error} is at most the modulus of continuity of the finitely many Jacobians on $U_0$, evaluated at $\|\Xi-\bar\Xi\|$, which tends to zero with its argument. Shrinking the ball until $M$ times this modulus is at most $\tfrac12$, and until its radius is below every $\bar\xi_i$ and $\bar\chi_i$ with $i\in\mathcal B$, gives $U$. With three times continuously differentiable predicates and model the Jacobians are Lipschitz continuous on $U_0$, and the modulus is at most a constant times $\|\Xi-\bar\Xi\|$.
\end{proof}

Let $\Sigma$ be the diagonal matrix with the entries $\bar\chi_i/\bar\xi_i$ for $i\in\mathcal B$ and zeros elsewhere. The nonsingularity holds when $DG(\bar\xi)$ has full row rank and, for every assignment in $\mathcal N(\bar\xi)$, the matrix $W^{\mathcal S}(\bar\xi,\bar\lambda)+\Sigma$ is positive definite on the null space of $DG(\bar\xi)$. When $\bar\xi$ lies in a support region, $\mathcal N(\bar\xi)$ holds the supports of $\bar\xi$ alone, and this is the usual second-order condition of the barrier problem. On the common boundary of two support regions that differ by one child at one node, as in \cref{sad:thm:switch}, it asks the condition of each one-sided Hessian and of none of their convex combinations.

\Cref{sad:thm:newton-switch} is local. Before the iterates reach its ball, the global convergence analysis of the line-search filter method for one barrier problem uses no second derivative of the objective or of the constraints \citep{wachter2002interior}. Its assumptions on the problem functions are values and first derivatives that are bounded and Lipschitz continuous on a set that holds the iterates. The sparsemax rows meet them on a bounded set with twice continuously differentiable predicates and model, because the projection onto the simplex is nonexpansive. Our solver declares at each restart the union of the old structure and the new active pattern (\cref{sad:subsec:declared}), so a solve restarts finitely often.

\subsection{Hessian structure declared to the solver}
\label{sad:subsec:declared}
IPOPT takes the structure of the Hessian at the start of a solve, and three declarations of it are possible. The static declaration of window cliques contains every active pattern and never needs a restart, but its size costs factorization time and the conversion of every declared entry at every call. An active declaration is small, and when a support leaves it the solve restarts from the current iterate on the union of the two structures, with the barrier parameter it had reached. A declaration of the supports' structure widened by a chosen number of samples in time, guarded by the runtime check of \cref{sad:subsec:reuse}, lies between the two. Our implementation builds it when that distance is positive.

\section{Additional Experimental Details}
\label{sad:app:evaluation}
\subsection{Sparsity patterns detected by CasADi}
\label{sad:subsec:app-casadi}
On a small instance with bounded windows, CasADi detects exactly the 138 prescribed entries of a Jacobian with 800 entries for both smoothings, although 103 to 115 of them are numerically zero for sparsemax at five test points. The instance has 40 time steps with two state coordinates each and ten constraints with always, eventually and until operators over windows of 4 to 8 samples. In CasADi the sparsemax threshold is found by bisection and then computed exactly on the support, and the values of both programs agree with our operators within $1.3\times10^{-15}$ at the five test points. CasADi colors the detected Jacobian with 18 forward or 4 reverse products and the detected Hessian pattern with 22 star colors, for both smoothings. On an instance of six always constraints that each read every column, CasADi detects all 480 entries for both smoothings, while the supports' pattern at two test points has 52 and 44 entries.

\begin{figure}[t]
\centering
\includegraphics{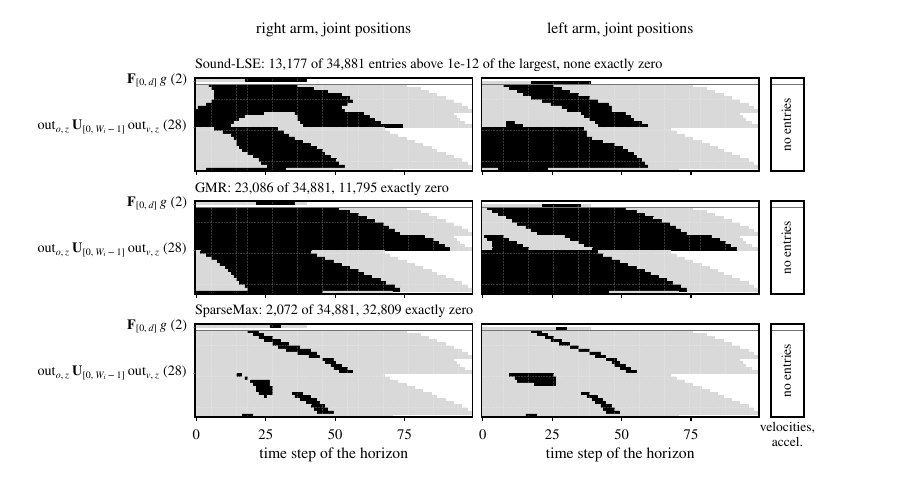}
\\[4pt]
\includegraphics{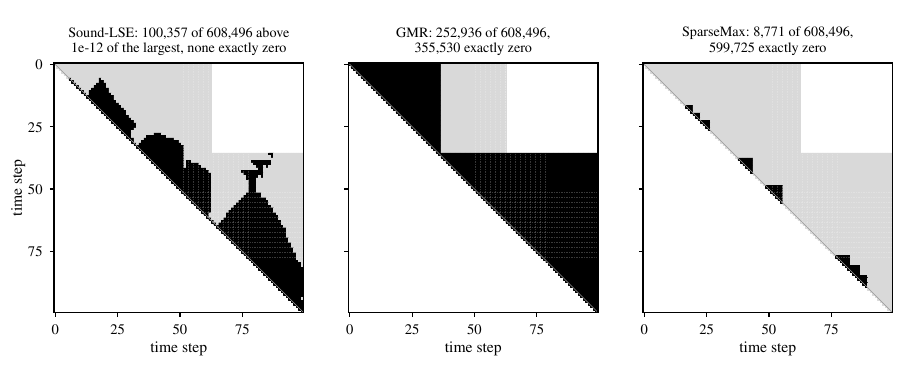}
\caption{Entries of the specification's constraints under \slse, \gmr{} and \smax{} at one control period of the reference motion with 100 time steps. Top: the rows of the constraint Jacobian on the 64-cell circle. Bottom: the Hessian on the five-cell zone, on the joint positions. Grey: the prescribed pattern. Black: entries above $10^{-12}$ of the largest.}
\label{sad:fig:three-smoothings}
\end{figure}

On the programs of \cref{sad:subsec:results} we compared the pattern that CasADi detects with the prescribed pattern entry by entry, at the control periods at which we time the derivatives. On the five-cell zone at 100, 200 and 400 time steps the two Jacobian patterns are equal, with 14{,}385, 31{,}199 and 60{,}221 entries. The detected Hessian pattern is the prescribed one together with 0, 84 and 686 diagonal entries of joint positions, which the tracking term of the cost contributes and which our prescribed pattern of the constraints does not hold. On the 64-cell circle the Jacobian patterns are equal, with 44{,}639, 184{,}639 and 452{,}039 entries, and so are the Hessian patterns. CasADi's own coloring of the detected Jacobian takes 7, 9 and 9 reverse products on the five-cell zone and 32, 98 and 130 on the 64-cell circle, where the greedy coloring of asdex takes 6, 8 and 8 and 31, 97 and 102. In 28 comparisons on six variants of the task at the same horizons, the Jacobian pattern detected for the sparsemax program had exactly the entries of the one detected for log-sum-exp, and both equalled the prescribed pattern. CasADi obtains the pattern of the Hessian from the expression of the gradient, without building the Hessian. \Cref{sad:fig:three-smoothings} shows the values behind this pattern at one control period on the 64-cell circle (top). CasADi detects the 34{,}881 prescribed entries of the specification's rows for log-sum-exp and for the GMR alike. Under log-sum-exp none of them is exactly zero, and 13{,}177 exceed $10^{-12}$ of the largest. Under the power mean robustness of order $(0,1)$, whose conjunction is the power mean of all operands when all are positive and of the negative parts otherwise, with the disjunction dual to it \citep{mehdipour2025generalized}, 11{,}795 are exactly zero. Under sparsemax the 2{,}072 nonzero entries are the supports' pattern, and the other 32{,}809 are exactly zero. The Hessian of the specification's part on the five-cell zone, with every multiplier equal to one, behaves alike (bottom). Of its 608{,}496 prescribed upper entries, log-sum-exp has none exactly zero and 100{,}357 above $10^{-12}$ of the largest, the GMR has 355{,}530 exactly zero, and sparsemax has 8{,}771 nonzero, the supports' pattern.

\subsection{Product counts and derivative times}
\label{sad:subsec:app-limits}
\begin{table}[t]
\centering
\caption{Derivative products and time of one evaluation, in ms unless a unit is given. Each cell lists dense differentiation in JAX, \slse{} in CasADi and \smax{} with our patterns, and a time cell has the generalized-mean robustness (GMR) in CasADi as its third entry. The product counts are medians over control periods, and the middle one is a greedy coloring of the prescribed pattern, which CasADi's own coloring exceeds by 1 to 28 (\cref{sad:subsec:app-casadi}). A dash marks a cell not run and a star one call.}
\label{sad:tab:counts}
\footnotesize
\setlength{\tabcolsep}{4pt}
\begin{tabular}{lcccc}
\toprule
 & \multicolumn{2}{c}{Constraint Jacobian} & \multicolumn{2}{c}{Hessian of the Lagrangian}\\
\cmidrule(lr){2-3}\cmidrule(lr){4-5}
$T$ & Products & Time & Products & Time\\
\midrule
\multicolumn{5}{l}{\emph{Five-cell zone}}\\
100 & 2{,}806 / 7 / 4 & 6.6~s / 2.6 / 2.7 / 2.7 & 4{,}200 / 1{,}011.5 / 59.5 & 26.6~s / 339 / 260 / 4.7\\
200 & 5{,}612 / 8 / 4 & 26.8~s / 5.8 / 5.8 / 3.7 & 8{,}400 / 1{,}204 / 63 & 109.6~s / 1{,}103 / -- / 6.9\\
400 & 11{,}222 / 8 / 4 & 114.9~s / 12.1 / 9.9 / 6.5 & 16{,}800 / 1{,}204 / 63 & 476.4~s / -- / -- / 17.7\\
\multicolumn{5}{l}{\emph{64-cell circle}}\\
100 & 2{,}823 / 25 / 9 & 3.9~s / 19.7 / 19.7 / 5.1 & 4{,}200 / 1{,}393 / 203 & 22.1~s / -- / -- / 11.2\\
200 & 5{,}713 / 95.5 / 13 & 27.4~s / 103.3 / 88.7 / 21.7 & 8{,}400 / 2{,}086 / 266 & 169.6~s / -- / -- / 53.2\\
400 & 11{,}493 / 102 / 13 & 137.2~s / 248.2 / 225.3 / 116.3 & 16{,}800 / 2{,}786 / 266 & 817.4~s$^{\ast}$ / -- / -- / 172.8\\
\bottomrule
\end{tabular}
\end{table}

\begin{figure}[t]
\centering
\includegraphics[width=\textwidth]{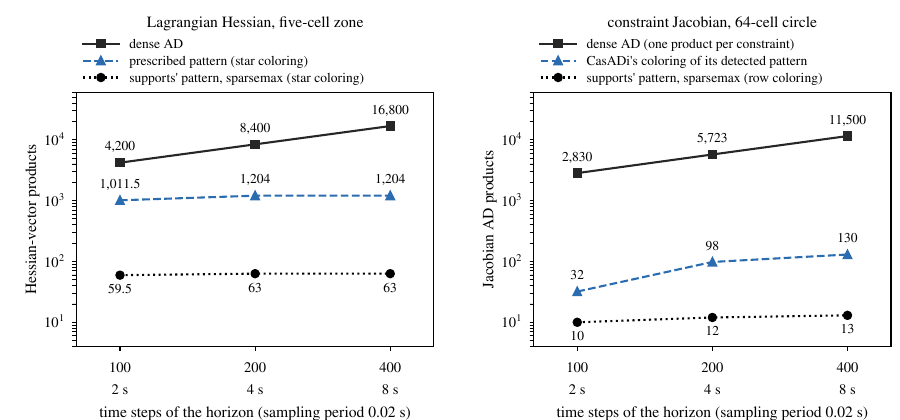}
\caption{Derivative products against the horizon at the control period of the derivative times, for the Hessian of the Lagrangian on the five-cell zone (left) and the constraint Jacobian on the 64-cell circle (right).}
\label{sad:fig:products-horizon}
\end{figure}

\Cref{sad:fig:products-horizon} plots the products against the horizon, on the five-cell zone as medians over ten control periods along the reference motion and on the 64-cell circle at the control period at which we time the derivatives, with CasADi's own coloring of the pattern it detects. The entries behave as the products do. On the five-cell zone at 200 time steps the Hessian has 35{,}284{,}200 upper entries, of which the prescribed pattern holds a median of 1.32 million and the supports' pattern 17{,}801. On the 64-cell circle at 200 time steps the Jacobian has 48{,}073{,}200 entries, of which the prescribed pattern holds 184{,}639 and the supports' pattern 27{,}762.

On the five-cell zone the prescribed count of the Hessian is the 14 joint positions of a time step times the 86 samples of the longest window, the largest number of variables that one until constraint reads, and it does not change between 200 and 400 time steps. The count with the supports is 63 at both horizons, so the support of no constraint covers more than 63 variables there.

On the 64-cell circle the prescribed count of the Jacobian is two more than the number of until windows that hold one time step, because every such constraint reads the joint positions of that time step and a row coloring gives each its own color. The supports' count is set by the cells whose hand-offs lie within the width of one support, about 11 samples, where a hand-off is the sample at which the visitor's rising distance to the cell meets the other arm's falling one. More cells raise both counts, and the lower bounds of their ratio at 200 time steps are 8.2, 8.6 and 8.8 with 64, 96 and 128 cells. A longer window raises the prescribed count alone, to 138 against 13 with windows of 150 samples at 400 time steps.

\subsection{Task, solver settings and additional solves}
\label{sad:subsec:app-solver}
The until conjunct of \cref{sad:eq:spec} is imposed one sample after the scheduled entry of the visitor. The reach conjunct is imposed at the passage of the previous waypoint, with the scheduled interval plus 10 samples as its deadline. The schedule is read from a reference motion on which neither arm rests and every conjunct holds in exact semantics with the margin of 0.02. A conjunct stays in the program until the recorded motion decides it, and the samples of its window that lie in the past enter as recorded values. The predicate $\mathrm{out}_{a,z}$ is a minimum node over one smooth predicate per sphere, the sphere's clearance to the cell divided by the cell's radius, and $g_{a,p}$ is normalized by its tolerance. Every node of a conjunct uses $\gamma=0.02~\mathrm{m}/L$ with $L$ the normalizing length of its predicates, 0.03~m for an until conjunct and 0.05~m for a reach conjunct, so that $\gamma$ is $2/3$ and $0.4$, and the log-sum-exp parameter follows from \cref{sad:eq:matching}. The conjunct counts of \cref{sad:subsec:setup} are those of the control period at which we time the derivatives, the one with the most until conjuncts at 100 time steps, and the product counts of \cref{sad:tab:counts} are medians over control periods. On both sides the reported solve is the second of its process, after a first one that includes the one-time compilation and setup.

\begin{figure}[t]
\centering
\includegraphics[width=\textwidth]{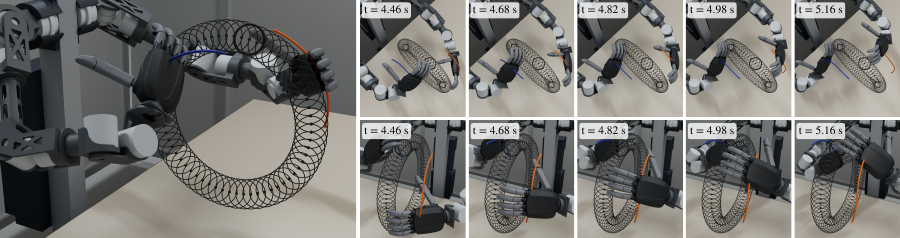}
\caption{The 64-cell circle in closed loop on the rigid-body simulation. Left: every cell as its outline and centre at 4.82~s. Right: the hand-off of one cell (thicker outline) from two viewpoints, where the right hand (blue path) leaves the cell before the left hand (orange path) enters it.}
\label{sad:fig:circle}
\end{figure}

\Cref{sad:fig:circle} shows the 64-cell circle in closed loop. Its cells are balls of radius 0.03~m whose centres lie 0.0157~m apart on a circle of radius 0.16~m.

Every time except those of \cref{sad:subsec:app-orin} comes from one workstation with a 16-core AMD Ryzen 9 7950X and 131~GB of memory, with one job at a time pinned to 12 cores. Our derivative evaluations run on 12 threads and on one thread, and CasADi's default evaluation of a function runs on one. Both sides call IPOPT 3.14.19 with the linear solver MUMPS~\citep{amestoy2001fully,amestoy2019performance} 5.8.2 on one thread through CasADi 3.8.1, with a tolerance of $10^{-6}$ on optimality and on the constraint violation and at most 1{,}000 updates of the iterate, or 3{,}000 in the hold-up scenario. The quasi-Newton approximation is limited-memory BFGS with a history of 10 updates. A solve from the previous control period's solution uses IPOPT's warm start, with its bound pushes and its initial barrier parameter at $10^{-6}$, on both sides. The shifted start holds the last state at zero velocity, shifts the multipliers of the defects and of the bounds by one sample, and carries the multiplier of every conjunct that is still imposed. Our derivatives are computed in JAX 0.11.2 with asdex 0.6.0. A derivative time is the median of 30 evaluations after compilation, and of 3 to 5 evaluations for dense differentiation, whose steps ran under a time budget that left one evaluation for the starred cell of \cref{sad:tab:counts}.

CasADi builds its exact Hessian of the log-sum-exp program on the five-cell zone in 349~s and 14~GiB at 100 time steps and in 1{,}345~s and 46~GiB at 200, and one evaluation of it takes 0.34~s and 1.10~s. At 400 time steps the solver with this Hessian takes 2{,}731~s and 88~GiB to build. The baseline's program is written in CasADi's scalar expressions.

\Cref{sad:tab:further-solves} gives the solves of \cref{sad:subsec:results} with other declared structures, and the cold solves of the hold-up scenario. Its two baselines are log-sum-exp and the generalized-mean robustness (GMR) in CasADi with IPOPT's quasi-Newton approximation, and with CasADi's exact Hessian the log-sum-exp baseline takes 12 updates and 5.50~s from the reference motion at 100 time steps and 15 updates and 29.25~s at 200. Widening the declared structure by 2 or 5 samples in time lowers the number of structures built in a cold solve and raises the cost of every Hessian. The supports' structure makes the solve from the previous period's solution 3.9 to 7.0 times shorter than the prescribed structure declared once, at the same 3 updates.

\begin{table}[t]
\centering
\caption{Further solves on the five-cell zone, as updates of the iterate and seconds, with the number of structures built in parentheses where it exceeds one. A hold-up is given in samples, a dash is a solve stopped at the update limit, and the times repeated in \cref{sad:tab:spread} are medians.}
\label{sad:tab:further-solves}
\footnotesize
\setlength{\tabcolsep}{4.5pt}
\begin{tabular}{llcccccc}
\toprule
Start & $T$ & \slse & \gmr & Supports & Widened by 2 & Widened by 5 & Prescribed\\
\midrule
Previous period & 100 & 67, 0.79 & 75, 0.95 & 3, 0.12 & 3, 0.14 & 3, 0.21 & 3, 0.48\\
 & 200 & 144, 5.53 & 135, 5.16 & 3, 0.22 & 3, 0.37 & 3, 0.48 & 3, 1.42\\
 & 400 & 142, 10.81 & 175, 14.48 & 3, 0.40 & 3, 0.48 & 3, 0.83 & 3, 2.81\\
Reference motion & 100 & 96, 1.18 & 126, 1.60 & 14, 0.77 (7) & 13, 0.59 (3) & 13, 0.82 (3) & 13, 1.95\\
 & 200 & 277, 11.36 & 228, 9.40 & 12, 1.27 (8) & 16, 1.42 (4) & 16, 2.36 (3) & 13, 10.37\\
 & 400 & 187, 14.09 & 228, 18.14 & 12, 2.06 (7) & 12, 1.78 (3) & 13, 2.40 (2) & 13, 21.49\\
Cold, hold-up 12 & 200 & 191, 7.02 & -- & 22, 1.61 (9) & 18, 1.24 (4) & 17, 2.01 (3) & 23, 20.95\\
Cold, hold-up 16 & 200 & 183, 5.37 & -- & 18, 1.48 (11) & 21, 1.73 (7) & 20, 2.32 (4) & 18, 12.30\\
Cold, hold-up 20 & 200 & 208, 6.55 & -- & 26, 2.67 (21) & 28, 2.58 (9) & 22, 2.14 (5) & 20, 11.55\\
\bottomrule
\end{tabular}
\end{table}

\Cref{sad:tab:spread} gives the spread of the times that we repeated, each run a new process alone on its cores. In \cref{sad:fig:replanning}b and c the quartiles of the 30 evaluations lie within 5 per cent of the median, and for our Jacobian on the 64-cell circle at 200 time steps between 21.3 and 25.8~ms around a median of 21.7~ms.

\begin{table}[t]
\centering
\caption{Spread of the repeated times in seconds, as the median with the smallest and the largest run. The solves are those of \cref{sad:tab:solves,sad:tab:orin}, and the closed loop is that of \cref{sad:fig:replanning}a at 100 time steps, with its slowest and its fastest control period. Every run takes the same updates.}
\label{sad:tab:spread}
\footnotesize
\begin{tabular}{llccc}
\toprule
 & $T$ & Runs & \slse, quasi-Newton & \smax, ours\\
\midrule
\multicolumn{5}{l}{\emph{Workstation}}\\
Solve, previous solution & 100 & 4 & 0.79 (0.77 to 0.82) & 0.122 (0.120 to 0.124)\\
 & 200 & 4 & 5.53 (5.45 to 5.62) & 0.225 (0.223 to 0.231)\\
 & 400 & 4 & 10.81 (10.70 to 10.87) & 0.403 (0.395 to 0.413)\\
Solve, after a 16-sample hold-up & 200 & 4 & 4.09 (4.01 to 4.11) & 0.218 (0.212 to 0.234)\\
Closed loop, slowest period & 100 & 6 & 1.40 (1.33 to 1.42) & 0.461 (0.459 to 0.469)\\
Closed loop, fastest period & 100 & 6 & 0.493 (0.490 to 0.504) & 0.092 (0.089 to 0.095)\\
\multicolumn{5}{l}{\emph{Jetson AGX Orin}}\\
Solve, previous solution & 100 & 4 & 2.38 (2.36 to 2.41) & 0.522 (0.515 to 0.543)\\
 & 200 & 4 & 13.68 (13.65 to 13.76) & 0.833 (0.801 to 0.859)\\
 & 400 & 4 & 28.49 (28.26 to 28.64) & 1.266 (1.231 to 1.350)\\
Solve, after a 16-sample hold-up & 200 & 4 & 11.04 (10.95 to 11.10) & 0.796 (0.776 to 0.823)\\
\bottomrule
\end{tabular}
\end{table}

\begin{table}[t]
\centering
\caption{Time of one evaluation of our derivatives in ms, on 12 threads and on one thread.}
\label{sad:tab:threads}
\small
\vskip 3pt
\begin{tabular}{llcc}
\toprule
Scene & $T$ & Jacobian, 12 threads / one & Hessian, 12 threads / one\\
\midrule
Five-cell zone & 100 & 2.7 / 4.9 & 4.7 / 10.4\\
 & 200 & 3.7 / 7.8 & 6.9 / 15.1\\
 & 400 & 6.5 / 14.1 & 17.7 / 39.4\\
64-cell circle & 100 & 5.1 / 6.4 & 11.2 / 19.2\\
 & 200 & 21.7 / 28.3 & 53.2 / 123.0\\
 & 400 & 116.3 / 132.2 & 172.8 / 355.0\\
\bottomrule
\end{tabular}
\end{table}

On the 64-cell circle at 200 time steps, from the previous control period's solution, the baseline takes 144 updates and 31.39~s and our solve on the supports' structure 3 updates and 5.67~s, of which 2.17~s declare the structure. With the prescribed structure it takes 3.66~s. Cold from the reference motion the baseline takes 284 updates and 59.33~s, our solve on the supports' structure 18 updates, 15 structures and 37.08~s, and on the prescribed structure 28 updates and 80.68~s. With the linear solver on 12 threads on both sides, the solves on the five-cell zone at 200 time steps take 5.56 and 11.41~s for the baseline and 0.25 and 1.28~s for ours, close to the times on one thread.

The closed loop of \cref{sad:subsec:results} runs on a rigid-body simulation of the two arms in MuJoCo \citep{todorov2012mujoco}, in which the joint drives track the plan between two solves. The controller starts at sample 100 of the motion delayed by a hold-up of 14 samples, re-plans every 25 samples over 100 time steps from the simulated joint positions and velocities and runs for 21 control periods, with the simulation waiting while it computes. The executed joint positions differ from the planned ones by at most 0.0040~rad. The collision spheres of the two arms stay at least 38~mm apart on the executed motion and overlap by 9~mm on the motion that ignores the hold-up, and the same constraints hold on the executed motion with either planner. On the 64-cell circle the same loop, with the declared structure widened by 2 samples, starts after a hold-up of 38 samples and runs for 13 control periods (\cref{sad:fig:circle}). All 225 until constraints and all 15 reach constraints decided during the run hold in exact semantics on the executed motion, the until constraints with a smallest robustness of 0.25 against $-0.22$ on the motion that ignores the hold-up. Every control period after the first is 1.7 to 3.1 times shorter than the baseline's solve alone.

After a hold-up the factor grows with the horizon, as it does from the previous solution. The median solve of the ten control periods after it is 6.3, 19 and 31 times shorter than the baseline's at 100, 200 and 400 time steps, and the first solve after it 1.7 to 4.4 times. A change of IPOPT's tolerance on both sides keeps the gain. At 200 time steps from the previous solution our solve is 15 times shorter with a tolerance of $10^{-4}$ and 44 times with IPOPT's default of $10^{-8}$, where the baseline stops at IPOPT's acceptable level after 231 updates.

\Cref{sad:tab:replanning} lists every re-planning loop that we timed on the planning model, over horizons of 2, 4 and 8~s with a new plan every 0.5~s and at every sample.

\begin{table}[t]
\centering
\caption{Re-planning loops on the planning model after a hold-up: time per control period in s, as the median and range of the ten periods after the first solve. The column on construction adds the baseline's construction of its program to its solve, and the factor divides its solve alone by our whole period.}
\label{sad:tab:replanning}
\small
\begin{tabular}{llcccc}
\toprule
Horizon & New plan every & Ours, whole period & Baseline, solve alone & With construction & Factor\\
\midrule
2~s & 0.5~s & 0.16 (0.11 to 0.40) & 0.68 (0.46 to 1.35) & 4.15 & 4.3\\
4~s & 0.5~s & 0.36 (0.25 to 1.76) & 5.99 (4.04 to 8.84) & 14.64 & 16.6\\
8~s & 0.5~s & 0.55 (0.50 to 2.23) & 14.06 (11.91 to 20.55) & 29.87 & 25.3\\
2~s & 0.02~s & 0.14 (0.11 to 0.23) & 0.98 (0.72 to 1.18) & 6.41 & 7.0\\
4~s & 0.02~s & 0.24 (0.20 to 0.35) & 4.18 (3.60 to 5.41) & 13.06 & 17.6\\
8~s & 0.02~s & 0.47 (0.43 to 0.94) & 13.18 (11.27 to 20.04) & 30.78 & 28.0\\
\bottomrule
\end{tabular}
\end{table}

\subsection{Comparison with the generalized-mean robustness}
\label{sad:subsec:app-gmr}
The baseline's second smoothing is the generalized-mean robustness \citep{mehdipour2025generalized} in its power mean form of order $(p,q)=(0,1)$, one of the three settings of its authors' control example. The measure is built so that all subformulas and time points contribute to the score, which removes the locality and masking of the traditional robustness. It is sound in its authors' sense for every order, which means that its sign is the sign of the exact robustness. Its conjunction is the power mean of exponent $p$ of the operands when all of them are positive \citep[Eq.~12]{mehdipour2025generalized}. Otherwise it is the negated power mean of exponent $q$ of the magnitudes of their negative parts, taken over all operands, so a positive operand enters as zero. Its disjunction follows by De Morgan's law, and the conjunction lies at or above the minimum. A constraint on the measure at the margin therefore certifies a positive exact robustness, but not in general the margin. Its authors note that with the limiting orders, as with the traditional robustness, the measure considers only the most critical time or subformula and is useful for monitoring or control in safety-critical applications \citep[p.~1954]{mehdipour2025generalized}. The other orders set the level of contribution of all subformulas and times. Sparsemax lies between the two. Its derivative stays on the operands within a threshold of the extremum, which one parameter sets, and its value bounds the exact robustness from below, so a margin on it carries over to the exact robustness.

We place the measure in each constraint of our program, in CasADi with its detection and IPOPT, as we do with log-sum-exp. We evaluate its authors' equations as printed, with the until nested as they define it \citep[Eq.~15]{mehdipour2025generalized} and without the smooth approximation of the positive and negative parts that they use to improve convergence. At our solutions of the solves started from the previous control period's solution, the derivative of the specification's constraints is then nonzero in 3{,}514 of the 4{,}627 Jacobian entries prescribed for them on the five-cell zone at 100 time steps. The counts are 8{,}491 of 11{,}641 at 200 time steps, 15{,}351 of 21{,}063 at 400 and 102{,}116 of 165{,}081 on the 64-cell circle at 200. The sparsemax supports hold 385, 679, 1{,}120 and 8{,}204 of the prescribed entries. We compared CasADi's detected patterns on five programs, each at one control period (the five-cell zone at the three horizons, the 64-cell circle and the hold-up scenario at 200 time steps). On each it detects for this measure the pattern it detects for log-sum-exp, entry for entry, in the Jacobian and in the Hessian, so at these control periods its derivative products equal those of log-sum-exp (\cref{sad:tab:counts}). CasADi's detection thus returns the prescribed Jacobian pattern on these programs for a second smoothing whose derivative, evaluated as printed, has exact zeros. One Jacobian of this program takes CasADi 0.8 to 1.0 times its log-sum-exp time.

From the previous control period's solution, where every conjunct holds, the baseline with this measure and IPOPT's quasi-Newton approximation takes 75, 135 and 175 updates and 0.95, 5.16 and 14.48~s at 100, 200 and 400 time steps (\cref{sad:tab:solves}). With log-sum-exp it takes 67, 144 and 142 updates. From the reference motion it takes 126, 228 and 228 updates, where log-sum-exp takes 96, 277 and 187 (\cref{sad:tab:further-solves}). In the hold-up scenario the first solve starts from the plan made before the hold-up, on which the delayed leading arm leaves 6 to 7 conjuncts violated. From that start, at hold-ups of 12, 16 and 20 samples and 200 time steps, IPOPT with its quasi-Newton approximation stopped at its limit of 3{,}000 updates. The tables therefore give no hold-up solve for this measure. Because the measure is not smooth at the satisfaction boundaries, its authors state that algorithms for nonsmooth optimization can be used and name two further aids, a first solve with a smooth measure and smooth approximations of the positive and negative parts \citep[Section~V]{mehdipour2025generalized}. We applied none of these.

Its authors maximize the measure in the cost under a margin constraint \citep[Eq.~11]{mehdipour2025generalized}, where we only constrain it. We also solved their formulation on our plant and specification at 100 time steps, with one constraint on the measure of the conjunction of all conjuncts, the weight of 100 of their control example and the margin of 0.02. IPOPT with the same quasi-Newton setting reached its limit of 1{,}000 updates from the reference motion and from the previous period's last iterate, at iterates on which every conjunct holds in exact semantics. Their own solves of that formulation use BFGS and SLSQP.

\subsection{Results on an embedded computer}
\label{sad:subsec:app-orin}
We repeated the solves and the derivative times of \cref{sad:subsec:results} on the 12-core CPU of a Jetson AGX Orin, with the same programs, options and software versions and no GPU (\cref{sad:tab:orin}). Our solves and the baseline's solve with CasADi's exact Hessian, 3 updates and 2.54~s at 100 time steps, reproduce the workstation's solutions to rounding. The quasi-Newton baseline takes a different number of updates there and stops at points that differ from the workstation's by up to $1.4\times10^{-3}$ in the solution. At 400 time steps IPOPT ends it with an error in the step computation, at a feasible iterate that misses the optimality tolerance and whose objective differs from the workstation's by $6\times10^{-7}$. On the device our solves are 4.6 to 22.5 times shorter than the baseline's, and our Hessian on the five-cell zone at 100 time steps is 29 times shorter than CasADi's (\cref{sad:tab:orin}). On the five-cell zone over four fifths of our Jacobian's time goes to the predicates and their derivatives, which every method has to evaluate, and on one thread the device's cores run them 2.5 to 3.1 times slower than the workstation's in our compiled code, against 1.7 to 2.1 times for CasADi's evaluation.

\begin{table}[t]
\centering
\caption{The comparison on the CPU of a Jetson AGX Orin: one solve as updates of the iterate and seconds, the baseline's with the quasi-Newton approximation / CasADi's exact Hessian, and one derivative evaluation in ms, ours on 12 threads / one thread. The solve's factor divides the baseline's shorter time by ours. The starred solve ended with an error. The quasi-Newton and our solve times are medians of four runs (\cref{sad:tab:spread}).}
\label{sad:tab:orin}
\small
\setlength{\tabcolsep}{5pt}
\begin{tabular}{llccc}
\toprule
 & $T$ & \slse{} in CasADi & \smax, ours & Factor\\
\midrule
Solve, previous solution & 100 & 74, 2.38 / 3, 2.54 & 3, 0.52 & 4.6\\
 & 200 & 137, 13.68 / 3, 12.35 & 3, 0.83 & 14.8\\
 & 400 & 126, 28.49$^{\ast}$ & 3, 1.27 & 22.5\\
Solve, after a 16-sample hold-up & 200 & 101 to 161, 11.04 & 3 to 7, 0.80 & 13.9\\
Hessian, five-cell zone & 100 & 630 & 21.9 / 26.6 & 29\\
Jacobian, five-cell zone & 100 & 4.5 & 12.3 / 14.0 & 0.36\\
 & 200 & 12.1 & 18.8 / 22.8 & 0.64\\
 & 400 & 23.3 & 27.5 / 39.0 & 0.85\\
Jacobian, 64-cell circle & 100 & 39.7 & 23.5 / 21.7 & 1.7\\
 & 200 & 206.6 & 76.0 / 99.9 & 2.7\\
 & 400 & 520.4 & 156.3 / 256.6 & 3.3\\
\bottomrule
\end{tabular}
\end{table}

%% file: figures/nnf-pattern-tree.tex
\begin{figure}[t]
\centering
\begin{tikzpicture}[x=1cm,y=1cm, every node/.style={font=\small},
  op/.style={draw,rounded corners=2pt,inner sep=4pt,fill=white},
  active/.style={->,thick,blue!65!black},
  inactive/.style={->,dashed,gray},
  edge/.style={->,black!65}]
\node at (0,3.4) {(a) syntax tree};
\node[op] (and) at (0,2.55) {\(\wedge\)};
\node[op] (eventually) at (-1,1.2) {\(\mathbf F_{[1,2]}\)};
\node[op] (always) at (1,1.2) {\(\mathbf G_{[0,1]}\)};
\node (mu) at (-1,0) {\(\mu\)};
\node (nu) at (1,0) {\(\neg\nu\)};
\draw[edge] (and) to (eventually);
\draw[edge] (and) to (always);
\draw[edge] (eventually) to (mu);
\draw[edge] (always) to (nu);
\node at (7.45,3.4) {(b) evaluation at \(t=1\) and current derivative paths};
\node[op] (root) at (7.45,2.55) {\(y_\varphi=Q^-_1(y_F,y_G)=63/64\)};
\node[op] (max) at (4.9,1.2) {\(y_F=Q^+_1(\cdot)=7/4\)};
\node[op] (min) at (10,1.2) {\(y_G=Q^-_1(\cdot)=1\)};
\node (mu2) at (3.5,0) {\(g_\mu(x_2)=2\)};
\node (mu3) at (6.3,0) {\(g_\mu(x_3)=1/2\)};
\node (nu1) at (8.6,0) {\(-g_\nu(x_1)=1\)};
\node (nu2) at (11.4,0) {\(-g_\nu(x_2)=3\)};
\draw[active] (root) to node[midway,fill=white,inner sep=1pt] {\(1/8\)} (max);
\draw[active] (root) to node[midway,fill=white,inner sep=1pt] {\(7/8\)} (min);
\draw[active] (max) to node[midway,fill=white,inner sep=1pt] {\(1\)} (mu2);
\draw[inactive] (max) to node[midway,fill=white,inner sep=1pt] {\(0\)} (mu3);
\draw[active] (min) to node[midway,fill=white,inner sep=1pt] {\(1\)} (nu1);
\draw[inactive] (min) to node[midway,fill=white,inner sep=1pt] {\(0\)} (nu2);
\end{tikzpicture}
\caption{The worked example of \cref{sad:subsec:worked}: (a) the syntax tree of \cref{sad:eq:nnf-example} and (b) its four leaf evaluations at $t=1$, all in the static graph. Solid edges carry a positive weight at the trajectory of \cref{sad:eq:example-states}, dashed edges a weight of zero, and arrows point from parents to children.}
\label{sad:fig:parse-tree}
\end{figure}